\documentclass[10pt,twocolumn,letterpaper]{article}

\usepackage{cvpr}              

\usepackage[dvipsnames]{xcolor}

\usepackage{graphicx}
\usepackage{subcaption}
\usepackage{bm}
\usepackage{booktabs}
\usepackage{siunitx}
\usepackage{colortbl, color}

\usepackage{siunitx}
\usepackage{comment}
\usepackage{listings}
\usepackage{amsmath, amssymb,enumitem}
\usepackage{times}
\usepackage{graphicx}
\usepackage{booktabs}
\usepackage{comment}
\usepackage{cases}
\usepackage{setspace}
\usepackage{overpic}
\usepackage{rotating}
\usepackage{subeqnarray}
\usepackage{multirow}
\usepackage{tabularx}
\usepackage{epstopdf}
\usepackage{tikz}
\usepackage{enumerate}
\usepackage{here}
\usepackage{bm}
\usepackage{color}
\usepackage{xcolor}
\usepackage{colortbl,booktabs}
\usepackage{bbm}
\usepackage[ruled,linesnumbered]{algorithm2e}
\usepackage{setspace}
\usepackage{tabularx}
\usepackage{diagbox}
\usepackage{siunitx}
\definecolor{cvprblue}{rgb}{0.21,0.49,0.74}

\usepackage[accsupp]{axessibility}
\usepackage[pagebackref,breaklinks,colorlinks,citecolor=cvprblue]{hyperref}

\usepackage[capitalize]{cleveref}
\crefname{section}{Sec.}{Secs.}
\Crefname{section}{Section}{Sections}
\Crefname{table}{Table}{Tables}
\crefname{table}{Tab.}{Tabs.}

\def\confName{CVPR}
\def\confYear{2023}

\newcommand{\RNum}[1]{\uppercase\expandafter{\romannumeral #1\relax}}

\definecolor{yellow}{RGB}{255, 255, 204}
\definecolor{lred}{RGB}{255, 204, 204}
\definecolor{lgreen}{RGB}{204, 255, 204}
\definecolor{lblue}{rgb}{0.9,0.95,1}
\definecolor{purple}{RGB}{229, 204, 255}
\definecolor{gray}{RGB}{240, 240, 240}
\definecolor{lgray}{gray}{0.95}
\definecolor{lyellow}{rgb}{1,1,0.92}

\definecolor{gold}{RGB}{248, 214, 99}
\definecolor{silver}{gray}{0.9}
\definecolor{bronze}{RGB}{231, 188, 133}

\begin{document}

\title{Real-Time 4K Super-Resolution of Compressed AVIF Images.\\AIS 2024 Challenge Survey}

\author{Marcos V. Conde$^\dagger$ \and 
Zhijun Lei$^\dagger$ \and 
Wen Li$^\dagger$ \and
Cosmin Stejerean$^\dagger$ \and
Ioannis Katsavounidis$^\dagger$ \and 
Radu Timofte$^\dagger$ \and
Kihwan Yoon \and
Ganzorig Gankhuyag \and
Jiangtao Lv \and Long Sun \and Jinshan Pan \and Jiangxin Dong \and Jinhui Tang \and
Zhiyuan Li \and Hao Wei\and Chenyang Ge \and
Dongyang Zhang \and 
Tianle Liu \and Huaian Chen \and Yi Jin \and
Menghan Zhou \and Yiqiang Yan \and
Si Gao \and
Biao Wu \and
Shaoli Liu \and
Chengjian Zheng \and
Diankai Zhang \and
Ning Wang \and
Xintao Qiu \and
Yuanbo Zhou \and
Kongxian Wu \and
Xinwei Dai \and
Hui Tang \and
Wei Deng \and
Qingquan Gao \and
Tong Tong \and
Jae-Hyeon Lee \and Ui-Jin Choi \and
Min Yan \and
Xin Liu \and
Qian Wang \and
Xiaoqian Ye \and
Zhan Du \and
Tiansen Zhang \and
Long Peng \and Jiaming Guo \and Xin Di \and Bohao Liao \and Zhibo Du \and Peize Xia \and Renjing Pei \and Yang Wang \and Yang Cao \and Zhengjun Zha \and
Bingnan Han \and Hongyuan Yu \and Zhuoyuan Wu \and Cheng Wan \and Yuqing Liu \and Haodong Yu \and Jizhe Li \and Zhijuan Huang \and  Yuan Huang \and Yajun Zou \and Xianyu Guan \and Qi Jia \and Heng Zhang \and Xuanwu Yin \and Kunlong Zuo \and
Hyeon-Cheol Moon \and Tae-hyun Jeong \and Yoonmo Yang \and Jae-Gon Kim \and Jinwoo Jeong \and Sunjei Kim \and
}


\twocolumn[{
\vspace{-3em}
\maketitle
\begin{center}
\setlength{\tabcolsep}{0pt}
\begin{tabular}{c c c}
\includegraphics[width=0.25\linewidth]{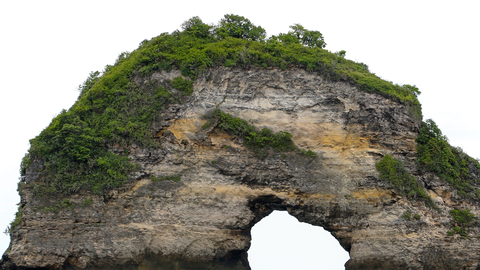} &
\includegraphics[width=0.25\linewidth]{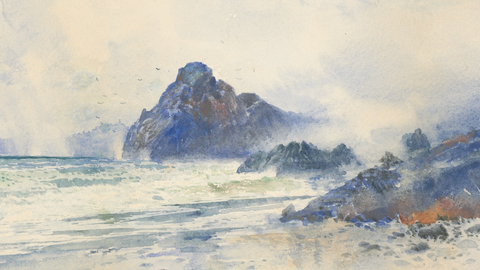} &
\includegraphics[width=0.25\linewidth]{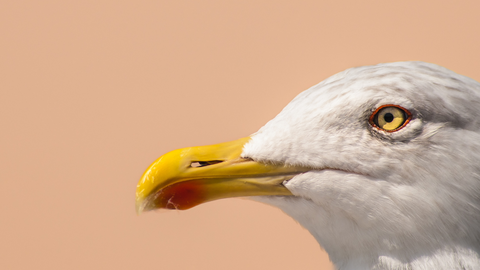} \\
\includegraphics[width=0.25\linewidth]{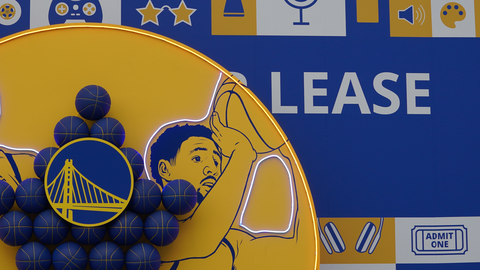} &
\includegraphics[width=0.25\linewidth]{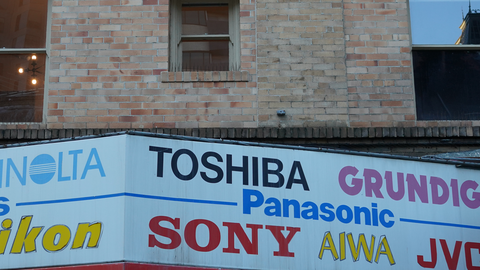} &
\includegraphics[width=0.25\linewidth]{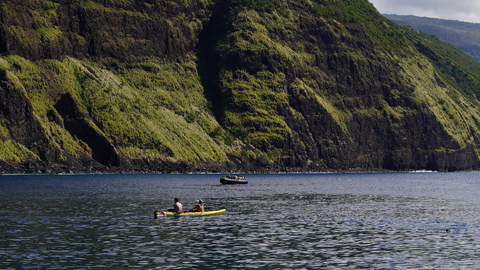} 
\end{tabular}
\captionof{figure}{Sample high-quality 4K images from the testing dataset of the \textbf{AIS 2024 RTSR Challenge.}}
\label{fig:samples}
\end{center}
\vspace{2em}
}]

\let\thefootnote\relax\footnotetext{$\dagger$ are the challenge organizers, while the other authors participated in the challenge. Workshop page: \url{https://ai4streaming-workshop.github.io/}. Benchmark code: \url{https://github.com/eduardzamfir/NTIRE23-RTSR}\\

\noindent Marcos V. Conde (corresponding author, project lead) and Radu Timofte are with University of W\"urzburg, CAIDAS \& IFI, Computer Vision Lab.\\
\noindent Zhijun Lei, Wen Li, Cosmin Stejerean and Ioannis Katsavounidis are with Meta.
}


\begin{abstract}
This paper introduces a novel benchmark as part of the AIS 2024 Real-Time Image Super-Resolution (RTSR) Challenge, which aims to upscale compressed images from 540p to 4K resolution (4x factor) in real-time on commercial GPUs. For this, we use a diverse test set containing a variety of 4K images ranging from digital art to gaming and photography. The images are compressed using the modern AVIF codec, instead of JPEG. All the proposed methods improve PSNR fidelity over Lanczos interpolation, and process images under 10ms. Out of the 160 participants, 25 teams submitted their code and models. The solutions present novel designs tailored for memory-efficiency and runtime on edge devices. This survey describes the best solutions for real-time SR of compressed high-resolution images.
\end{abstract}


\section{Introduction}
\label{rsec:introduction}
Single image super-resolution (SR) methods generate a high-resolution (HR) image from a single degraded low-resolution (LR) image. This ill-posed problem was initially solved using interpolation methods. However, SR is now commonly approached through the use of deep learning~\cite{DIV2K, espcn, wang2018esrgan, liang2021swinir, swin2sr}.
Image SR assumes that the LR image is obtained through a degradation processes. This can be expressed as:

\begin{equation}
\mathbf{y} = (\mathbf{x} * \mathbf{k})\downarrow_s,
\end{equation}

where $\mathbf{*}$ represents the convolution operation between the LR image and the blur kernel, and $\downarrow_s$ is the down-sampling operation with respective down-sampling factor $\times s$ (\eg $\times 2$, $\times 3$, $\times 4$, $\times 8$).

Following the foundational efforts by Shi \etal~\cite{shi2016real}, optimizing deep neural networks for single image super-resolution has become critical~\cite{zhao2020efficient,wang2021exploring,song2020efficient,hui2019imdn,kong2022residual, zamfir2023rtsr}. This focus has inspired the creation of numerous workshops and challenges, for instance \cite{zhang2020aim, li2022ntire}, which serve as platforms for exchanging ideas and pushing the boundaries of efficient and real-time super-resolution (SR). Moreover, many works consider compressed images as inputs (\eg JPEG), what leads to more practical super-resolution methods~\cite{yang2022aim, conde2023efficient, swin2sr}.


\begin{table*}[t]
    \centering
    \resizebox{\linewidth}{!}{
    \begin{tabular}{r c c c c c c c c c c c c c}
        \toprule
        \rowcolor{lgray} Team Method & Score & $\Delta$ & MACs & Runtime & \# Par. & \multicolumn{2}{c}{PSNR-RGB [dB]} & \multicolumn{2}{c}{PSNR-Y [dB]} & \multicolumn{2}{c}{SSIM-RGB} & \multicolumn{2}{c}{SSIM-Y} \\
        
        \rowcolor{lgray} &  & [dB] & [G] & [ms] & [M] &  QP31 & QP63 & QP31 & QP63 & QP31 & QP63 & QP31 & QP63 \\
        \midrule
        
        BasicVison (\cref{sec:basicvision}) & 27.27 & 0.350 &  6.20 &  0.874 &  0.012 & 30.85 & 26.83 & 33.30 & 29.27 & 0.807 & 0.719 & 0.850 & 0.777 \\
        
        CMVG (\cref{sec:basicvision}) & 27.55 & 0.330 &  5.16 & 0.833 &  0.010 & 30.84 & 26.82 & 33.27 & 29.26 & 0.807 & 0.719 & 0.849 & 0.777 \\
        
        IVP (\cref{sec:basicvision}) & 27.08 & 0.365 &  6.07 & 0.905 &  0.011 & 30.85 & 26.83 & 33.33 & 29.27 & 0.808 & 0.719 & 0.851 & 0.777 \\
        
        RVSR (\cref{sec:xjtu}) & 12.00 & 0.720 &  15.6 & 7.542 &  0.033 & 31.52 & 27.01 & \cellcolor{silver} 33.88 & \cellcolor{silver}29.43 & 0.820 & 0.725 & 0.859 & 0.781 \\
        
        VPEG-R (\cref{sec:vpeg}) & 25.77 & 0.100 & \cellcolor{silver} 1.56 & 0.692 & 0.012 & 30.59 & 26.70 & 32.94 & 29.13 & 0.803 & 0.714 & 0.845 & 0.773 \\
        
        VPEG-S (\cref{sec:vpeg}) & 16.15 & 0.735 & 34.2 & 4.251 & 0.066 & 31.57 & 26.99 & \cellcolor{gold} 33.93 & \cellcolor{gold} 29.41 & 0.821 & 0.725 & 0.859 & 0.781 \\
        
        ANUNet (\cref{sec:lenovo}) & 22.77 & 0.545 & 9.40 & 1.642 &  0.072 & 31.27 & 26.86 & 33.66 & 29.30 & 0.814 & 0.719 & 0.855 & 0.778 \\
        
        RESR (\cref{sec:FZUQXT}) & 28.68 & 0.455 & 7.04 & 0.914 &  0.044 & 31.16 & 26.87 & 33.50 & 29.28 & 0.813 & 0.720 & 0.853 & 0.777 \\
        
        Megastudy (\cref{sec:megastudy}) & 13.12 & 0.210 & 20.7 & 3.109 &  0.040 & 30.76 & 26.87 & 33.01 & 29.28 & 0.801 & 0.721 & 0.842 & 0.778 \\
        
        CASR (\cref{sec:CASR_z6})  & \cellcolor{gold} 33.70 & 0.205 & \cellcolor{bronze} 2.10 & \cellcolor{silver} 0.468 &  0.026 & 30.64 & 26.71 & 33.11 & 29.17 & 0.806 & 0.715 & 0.848 & 0.774 \\
        
        XiaomiMM (\cref{sec:xiaomi})  & 18.60 & 0.690 & 13.3 & 3.010 &  0.026 & 31.41 & 26.96 & \cellcolor{bronze}33.85 & \cellcolor{bronze}29.40 & 0.819 & 0.725 & 0.857 & 0.781 \\

        Team C3 (\cref{sec:xiaomi}) & 20.35 & 0.500 & 12.4 & 1.932 &  0.024 & 31.12 & 26.90 & 33.52 & 29.35 & 0.813 & 0.723 & 0.853 & 0.780 \\

        USTC Noah (\cref{sec:noah}) & 25.63 & 0.485 & 8.56 & 1.193 & 0.045 & 31.18 & 26.90 & 33.52 & 29.32 & 0.814 & 0.722 & 0.854 & 0.779 \\

        Lanczos++ (\cref{sec:zte}) & \cellcolor{bronze} 33.24 & 0.070 & \cellcolor{gold} 1.25 & \cellcolor{gold} 0.399 & 0.022 & 30.55 & 26.71 & 32.88 & 29.13 & 0.800 & 0.714 & 0.843 & 0.772 \\

        PixelArtAI (\cref{sec:smallnets}) & \cellcolor{silver} 33.32 & 0.395 & 9.68 &  \cellcolor{bronze} 0.623 & 0.075 & 31.06 & 26.84 & 33.40 & 29.26 & 0.811 & 0.719 & 0.852 & 0.777 \\
        
        RepTCN (\cref{sec:smallnets}) & 30.91 & 0.355 & 5.02 & 0.685 & 0.010 & 30.97 & 26.83 & 33.31 & 29.27 & 0.809 & 0.719 & 0.850 & 0.777 \\
        
        \midrule

        URPNet (\cref{sec:URPNet})  & - & - & 1.24 & 0.621 &  0.009 & 30.33 & 26.65 & 32.64 & 29.07 & 0.798 & 0.713 & 0.842 & 0.773 \\
        
        \rowcolor{lgray} Baseline Lanczos & - & - & - & - & - & 30.36 & 26.67 & 32.75 & 29.12 & 0.800 & 0.715 & 0.843 & 0.774 \\
        
        \bottomrule
    \end{tabular}
    }
    \caption{\textbf{Results of the AIS24 Real-Time SR challenge.} All the proposed methods upsample the images under \textbf{8ms}. The models can process compressed images with QP factors from 31 to 63. We provide PSNR and SSIM metrics in the RGB domain, and for the Luma (Y) channel. We report the fidelity improvement \emph{w.r.t} the baseline in terms of PSNR $\Delta$, considering the average PSNR-Y for QP31 and QP63. We highlight the top-3 (gold, silver, bronze) methods in terms of fidelity, runtime, and computational efficiency (MACs operations).
    }
    \label{tab:ais_benchmark}
\end{table*}

\section{AIS 2024 Real-Time Image SR Challenge}
\label{rsec:challenge}
In conjunction with the 2024 AIS: Vision, Graphics and AI for Streaming workshop, we introduce a new real-time 4K super-resolution challenge.

The challenge aims to upscale a compressed LR image from $540$p to 4K resolution using a neural network that complies with the following requirements: (i) improve performance over Lanczos interpolation. (ii) Upscale the image under 33ms. Moreover the images are compressed using different compression factors (QP values) using the modern AVIF codec instead of JPEG. The challenge seeks to identify innovative and advanced solutions for real-time super-resolution of compressed images.

\subsection{Motivation}

AV1 Image File Format (AVIF) is the latest royalty-free image coding format developed based on the Alliance for Open Media's (AOM) AV1 video coding standard. The compression efficiency and quality of AVIF encoded images is noticeably superior to JPEG and also HEIC, which uses HEVC for image coding. AVIF is also supported in all major web browsers. In the AIS 2024 Real-Time Image SR Challenge, we want to leverage AVIF as the image coding format to evaluate the quality improvement from SR when combining with AVIF. 

\subsection{4K SR Benchmark Dataset}

Following~\cite{conde2023efficient}, the \emph{4K RTSR benchmark} provides a unique test set comprising ultra-high resolution images from various sources, setting it apart from traditional super-resolution benchmarks. Specifically, the benchmark addresses the increasing demand for upsampling computer-generated content \eg gaming and rendered content, in addition to photo-realistic imagery, thereby posing a different challenge for existing SR approaches. 

The testing set includes diverse content such as rendered gaming content, digital art, as well as high-resolution photo-realistic images of animals, city scenes, and landscapes, totaling $110$ test samples. 

All the images in the benchmark testing set are at least 4K resolution \ie $3840\times2160$ (some are bigger, even 8K). 

The \textbf{distribution} of the 4K RTSR benchmark testset is: 14 real-world captures using a 60MP DSLR camera, 21 rendered images using Unreal Engine, 75 diverse images \eg animals, paintings, digital art, nature, buildings, etc.

\paragraph{Compression and Downsampling}
We use \texttt{ffmpeg} to produce the LR compressed images. We use 5 different QP values: 31, 39, 47, 55, 63. We use lanczos interpolation to downsample the images. Bellow we provide an example:

\lstset{
    language=bash,                       
    basicstyle=\small\ttfamily,          
    keywordstyle=\color{black},          
    commentstyle=\color{black},          
    stringstyle=\color{black},           
    showstringspaces=false,              
    breaklines=true,                     
    breakatwhitespace=false,             
    postbreak=\mbox{\textcolor{black}{$\hookrightarrow$}\space}, 
    tabsize=2,                           
    frame=single,                        
    numbers=none                         
}

\begin{lstlisting}
ffmpeg -hide_banner -y -loglevel error -i ../1.png -vf 'scale=ceil(iw/4):ceil(ih/4):flags=lanczos+accurate_rnd+full_chroma_int:sws_dither=none:param0=5' -c:v libsvtav1 -qp 31 -preset 5 1_4x_qp31.avif
\end{lstlisting}

In the context of AVIF and AV1 codecs, larger Quantization Parameter (QP) values imply more compression. Essentially, the QP value dictates the level of quantization applied to the video or image data, where higher quantization reduces the amount of data required to represent the original input, thus leading to higher compression ratios.

The participants can use any publicly available dataset, and produce the corresponding LR images.

\subsection{Evaluation}

The evaluation scripts were made available to the participants through GitHub (\url{https://github.com/eduardzamfir/NTIRE23-RTSR}). This allowed the participants to benchmark the performance of their models on their systems. During the final test phase, the participating teams provided the code, models and results corresponding to the $110$ test images. They did not have access to the HR ground-truth. The organizers then validated and executed the submitted code to obtain the final results, which were later conveyed to the participants upon completion of the challenge.


\subsection{Architectures and Main Ideas}
Here we summarize the core ideas behind the most competitive solutions. Note that most of the ideas follow~\cite{conde2023efficient, zamfir2023rtsr}.

\begin{itemize}
    \item \textbf{Re-parameterization} enables training the network using sophisticated blocks~\cite{ding2021repvgg}, while allowing these ``RepBlocks" to be simplified into a standard $3\times3$ convolutions during inference. This technique has become state-of-the-art in efficient SR~\cite{ntire22efficientsr, conde2023efficient}.
    
    \item \textbf{Pixel shuffle and unshuffle.}  These techniques are also known as depth-to-space, space-to-depth, and sub-pixel convolutions~\cite{shi2016real}. These are utilized to effectively apply spatial upsampling and downsampling over feature maps.
    
    \item \textbf{Multi-stage Training:} Given the significant limitations and shallow architecture of the neural networks, this approach enhances learning by varying learning rates and loss functions sequentially.
    
    \item \textbf{Knowledge distillation} allows to transfer knowledge from complex neural networks into more efficient ones.
    
\end{itemize}


\subsection{Results and Conclusion}

In \cref{tab:ais_benchmark} we provide the challenge benchmark. The models can upsample compressed 540p images and recover the core structural information according to the metrics calculated over Luma (Y). We can also appreciate a notable performance decay at high QP (compression) values. The runtime is the average of 100 runs after GPU warm up, using an NVIDIA 4090 GPU. Following previous work, we also report a ``Score" (S) that considers PSNR and runtime, defined as: $S = (2\times 2^{\Delta}) \mathbin{/} (T^{0.5} \times C$) where $T$ is the runtime, and $C=0.1$ is a constant scaling factor.

Attending to the results, we consider that this score is now obsolete, and we included it only for informative purposes. All the methods achieve runtimes under 10ms. In this scenario, larger fidelity improvements and other memory-related improvements are considered more important for practical applications.

In \cref{rsec:teams} we provide the description of the top solutions. 

Considering the best methods, we can conclude that there is certain convergence in the model designs. As previously mentioned, re-parameterization is ubiquitous. Edge-oriented filters to extract directly high-frequencies allow to reduce sparsity in the neural network, making effective use of all the kernels (parameters). Upsampling the input image, and enhancing it through a global residual connection is also a common neural network architecture.

\vspace{-2mm}
\paragraph{Related Challenges}
This challenge is one of the AIS 2024 Workshop associated challenges on: Event-based Eye-Tracking~\cite{wang2024ais_event}, Video Quality Assessment of user-generated content~\cite{conde2024ais_vqa}, Real-time compressed image super-resolution~\cite{conde2024ais_sr}, Mobile Video SR, and Depth Upscaling. 

\vspace{-2mm}
\paragraph{Acknowledgments}
This work was partially supported by the Humboldt Foundation. We thank the AIS 2024 sponsors: Meta Reality Labs, Meta, Netflix, Sony Interactive Entertainment (FTG), and the University of W\"urzburg (Computer Vision Lab).

Marcos Conde is also supported by Sony Interactive Entertainment (FTG).


\section{Methods and Teams}
\label{rsec:teams}

In the following sections we describe the best challenge solutions. Note that the method descriptions were provided by each team as their contribution to this survey. 

\begin{figure}[t]
    \centering
    \includegraphics[width=\linewidth]{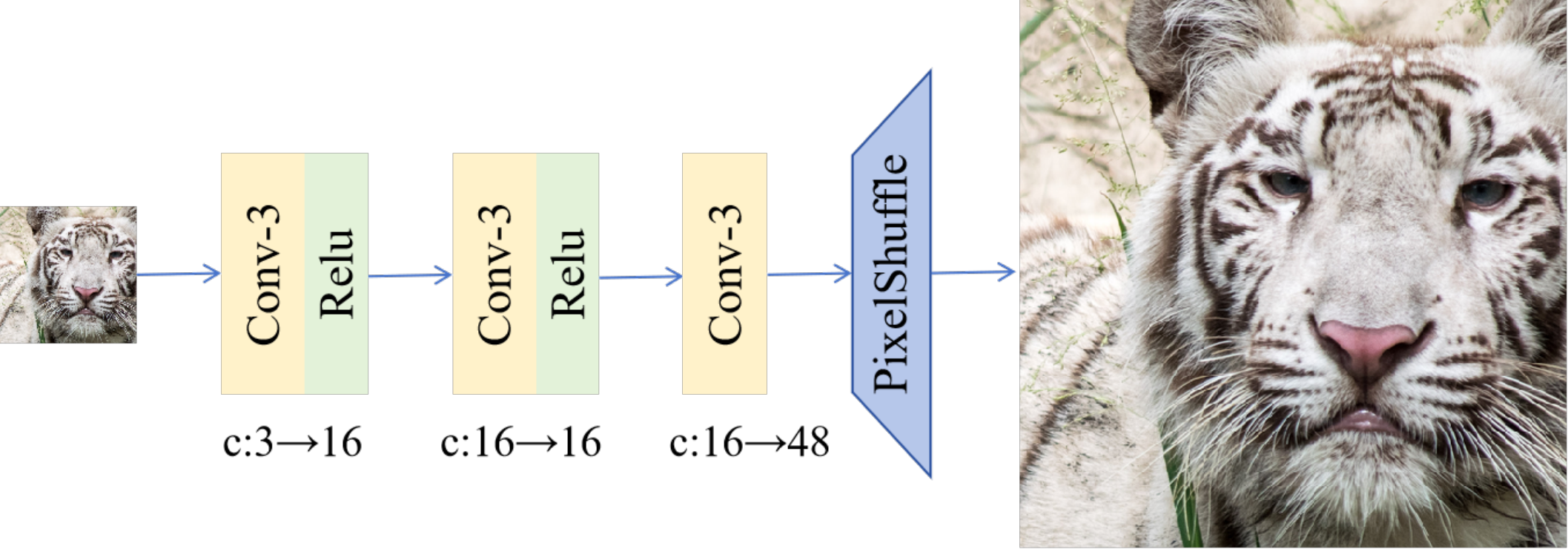}
    \caption{Overview of RepTCN by Team CameraAI.}
    \label{fig:RepTCN}
\end{figure}

\subsection{Simple Baselines}
\label{sec:smallnets}

\begin{center}

\vspace{2mm}
\noindent\emph{\textbf{Team CameraAI}}
\vspace{2mm}

\noindent\emph{
Tianle Liu,
Huaian Chen,
Yi Jin
}

\vspace{2mm}

\noindent\emph{
University of Science and Technology of China}

\end{center}


The team proposes \textbf{RepTCN}, a network comprising only three convolutional layers, achieving superior performance over Lanczos interpolation while maintaining exceptional efficiency. To further enhance efficacy, we introduced re-parameterization techniques, replacing the middle convolutional layer with a RepBlock \cite{repvgg} during the training phase. Additionally, we devised a three-stage training strategy to fully exploit the model's potential.

Figure \ref{fig:RepTCN} illustrates our proposed RepTCN. It consists of three convolutional layers, each without bias. A ReLU activation function is applied between every two convolutional layers. During the training phase, we replace the middle convolution with a RepBlock\cite{repblock}. During inference, we reparameterize the RepBlock into a convolutional layer. 

\paragraph{Implementation details}

Our training framework uses Pytorch for training on the RTX3090. We gathered the first 600 images from DIV2K, the first 600 images from Flicker2K, and the first 800 images from GTAV. Subsequently, we cropped these images to $512 \times 512$ to form our dataset.

During the training phase, the input from the dataset will be randomly cropped into patches, and these patches will undergo random horizontal flips and rotations. The model training can be divided into three stages. In the first stage, we set the batch size to 32 and the patch size to 32. L1 loss are used as target loss functions. We replaced the middle convolutional layer with a RepBlock\cite{repblock} and trained for 1000k iterations using the Adam optimizer, with a learning rate of $1 \times 10^{-3}$ decreasing to $1 \times 10^{-7}$ through the cosine scheduler. In the second stage, we set the batch size to 16 and the patch size to 128. MSE loss are used as target loss functions. We reparameterized the RepBlock into a convolutional layer and trained for 500k iterations using the Adam, with a learning rate of $5 \times 10^{-4}$ decreasing to $5 \times 10^{-7}$ through the cosine scheduler. In the third stage, we removed the bias from each convolutional layer and trained for 2000k iterations using the Adam, with a learning rate of $5 \times 10^{-4}$ decreasing to $5 \times 10^{-7}$ through the cosine scheduler. The other Settings are the same as in the previous step.

\begin{figure}[t]
  \centering
  \includegraphics[totalheight=2in]{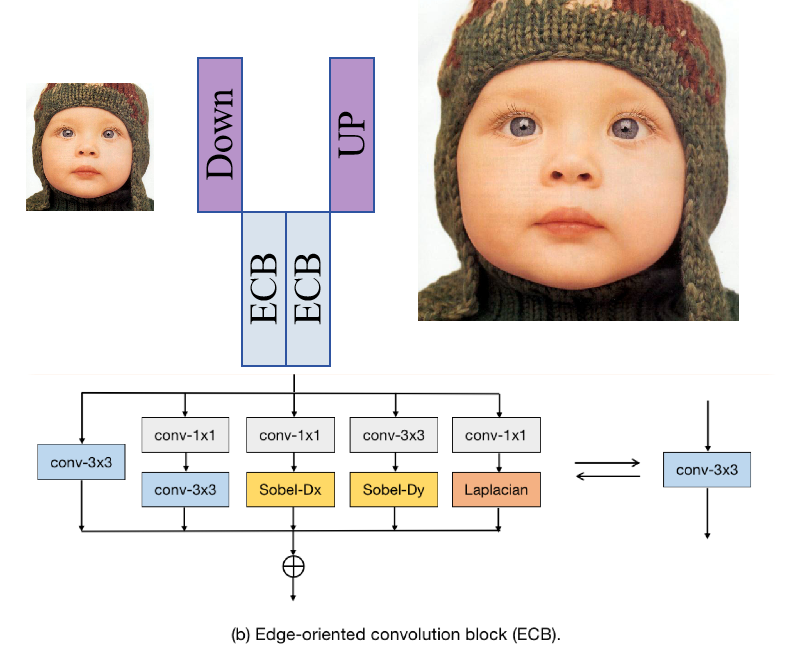}
  \caption{Simple network proposed by PixelArtAI.} \label{fig:PixelArtAI}
\end{figure}

\begin{center}

\vspace{2mm}
\noindent\emph{\textbf{Team PixelArtAI}}
\vspace{2mm}

\noindent\emph{
Dongyang Zhang
}

\vspace{2mm}

\noindent\emph{
MangoTV}

\end{center}

The team proposes a lightweight and extremely low-time-consuming network is built through re-parameters. Based on the ECB module~\cite{zhang2021edge}, we designed a lightweight and low-time-consuming network for the competition. The network design points are as follows:

First downsample by a factor of 2 using a convolution with a stride of 2. Downsampling breaks down compression and also improves network inference speed. Then stack two ECB modules and a 8x upsampled pixel shuffle module to return a three-channel image -- see \cref{fig:PixelArtAI}.

\vspace{-2mm}

\paragraph{Efficiency Metrics} Considering an input 540p and x4 SR, the model has 1.8798 KMACs and a runtime of 1.0367 ms.

\vspace{-2mm}

\paragraph{Implementation details}

The training data is degraded by FFmpeg with random QP. The input image size is 120x120x3, amd the batch size is 96. We use Adam optimizer with the initial learning rate set to 0.001. The training is divided into two stages: First, the learning rate is 0.001 and the loss is L1. This stage is trained for 60k iterations. Second, only the PSNR Loss is calculated, and the initial learning rate set to 0.0002, and is halved by 20k iterations.

\begin{figure*}
    \centering
    \includegraphics[width=0.95\linewidth]{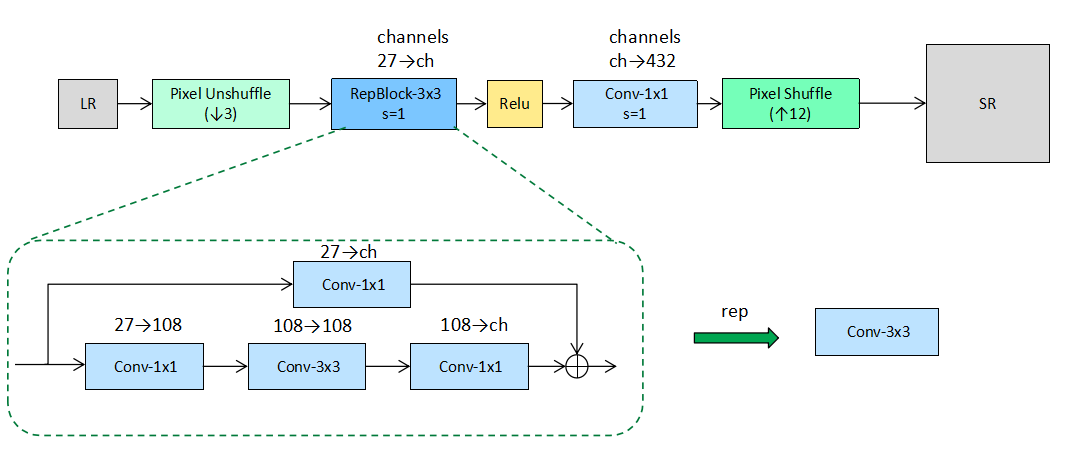}
    \caption{Lanczos++ proposed by Team ZXVIP}
    \label{fig:lanczos+}
\end{figure*}

\begin{table}
    \centering
    \begin{tabular}{l c}
         \toprule
         Method & PSNR  \\
         \midrule
         Lanczos & 22.70  \\
         Bicubic++ & 22.92  \\
         RCBSR & 22.76  \\
         Lanczos++  & 22.77 \\
         \bottomrule
    \end{tabular}
    \caption{Lanczos++ compared with other fast methods on the 40 images validation set.}
    \label{tab:lanczos+}
\end{table}


\subsection{Lanczos++:An ultra lightweight image super-resolution network}
\label{sec:zte}

\begin{center}

\vspace{2mm}
\noindent\emph{\textbf{Team ZXVIP}}
\vspace{2mm}

\noindent\emph{
Si Gao ~$^1$,
Biao Wu ,
Shaoli Liu,
Chengjian Zheng,
Diankai Zhang,
Ning Wang ~$^2$
}

\vspace{2mm}

\noindent\emph{
$^1$ Network Intelligence Platform Development Dept. III, ZTE\\
$^2$ Network Intelligence Platform System Dept., ZTE}

\end{center}


The team proposes an ultra lightweight image super-resolution network named Lanczos++.

The highlights of the proposed network (see \cref{fig:lanczos+}) are as follows: First, we use PiexlShuffle to perform 3x down-sampling on the input LR image while increasing the channel dimension. This design significantly improves the inference efficiency of the network, while basically not losing the representation ability of the model. Second, we design a new type of reparameterization module named Rep-Block(see \cref{fig:lanczos+}). 

In the training phase, we first use 1x1 convolution to increase the input channel to four times, then use 3x3 convolution for feature extraction, and finally use 1x1 convolution to transform the dimension into the output channel dimension, and use it as a residual with the branch that transforms the input channel into the output channel through 1x1 convolution. During the model inference stage, we merge the reparameterization module into a standard 3x3 convolution. Reparameterization can improve the fidelity of the model while maintaining its inference efficiency unchanged. 

Third, we remove the bias of the convolutional layer and use 1x1 convolution instead of 3x3 convolution in the reconstruction layer convolution section, which can significantly reduce the running time of the model. Finally, we use Pixelshuffle for 12x up-sampling.

\vspace{-2mm}

\paragraph{Implementation details}

We use 4450 images from DIV2K, Flickr2K and GTA V datasets for training.To generate the LR data, the images are lanczos downsampled by scale 4 and compressed samples using AVIF with QP 31/39/47/55/63 (5 compression levels).

We implement a three-stage training pipeline: training a basic model, deviation removal of convolutional layers, and final fine-tuning of switching loss functions. In the first stage, we conduct a NAS architecture search to find the optimal network parameter configuration. For the first two stages, we use the L1 loss function for training, and for the final stage, we use the L2 loss function. 

We use Adam optimizer by setting  $\beta_1 = 0.9$  and $\beta_2 = 0.999$. In the first two stages of training, we start with a learning rate of 5e-4. For the final stage, they start from 2e-4. We use a decaying learning rate scheduler for all stages, where the first 500 epochs are preheated and then the learning rate decays linearly until 1e-8. The total duration of the training process is around 50h using a V100 (32Gb) GPU.


\begin{figure*}[t] 
\centering
\begin{minipage}[t]{0.5\linewidth}
    \includegraphics[width=\textwidth,height=0.3\textwidth]{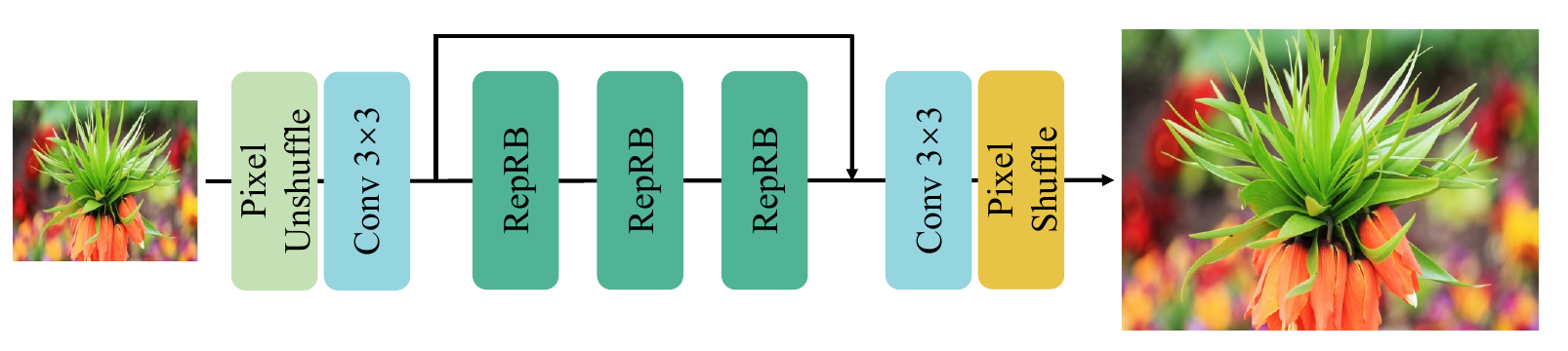}
    \caption{An overview of the proposed VPEG-R model.} 
    \label{fig:vpegr}
\end{minipage}\hfill
\begin{minipage}[t]{0.4\linewidth}
    \includegraphics[width=\textwidth,height=0.3\textwidth]{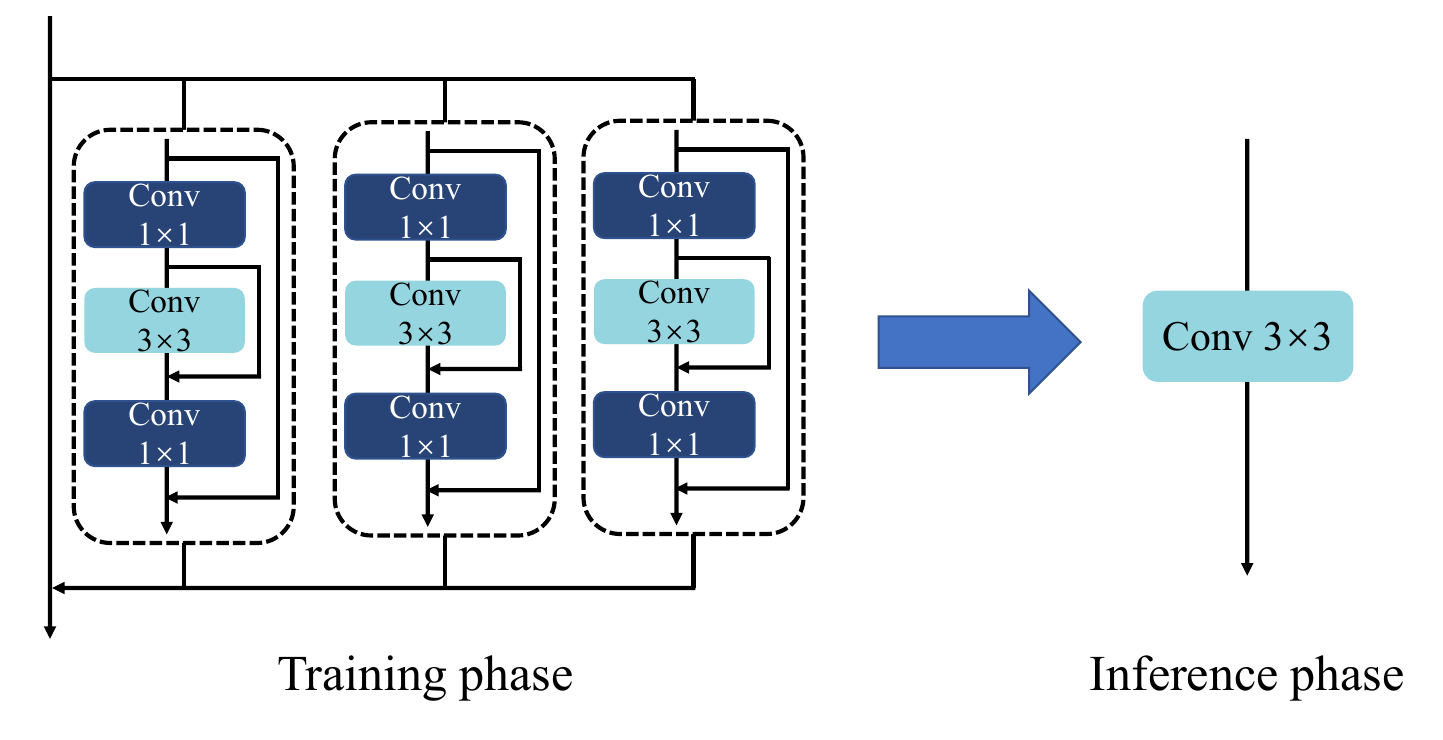}
    \caption{Proposed RepRB for the VPEG-R model.} 
    \label{fig:reprb}
\end{minipage}
\label{fig:vpegr_block}
\end{figure*}

\begin{figure*}[t]
    \centering
    \includegraphics[width=\textwidth]{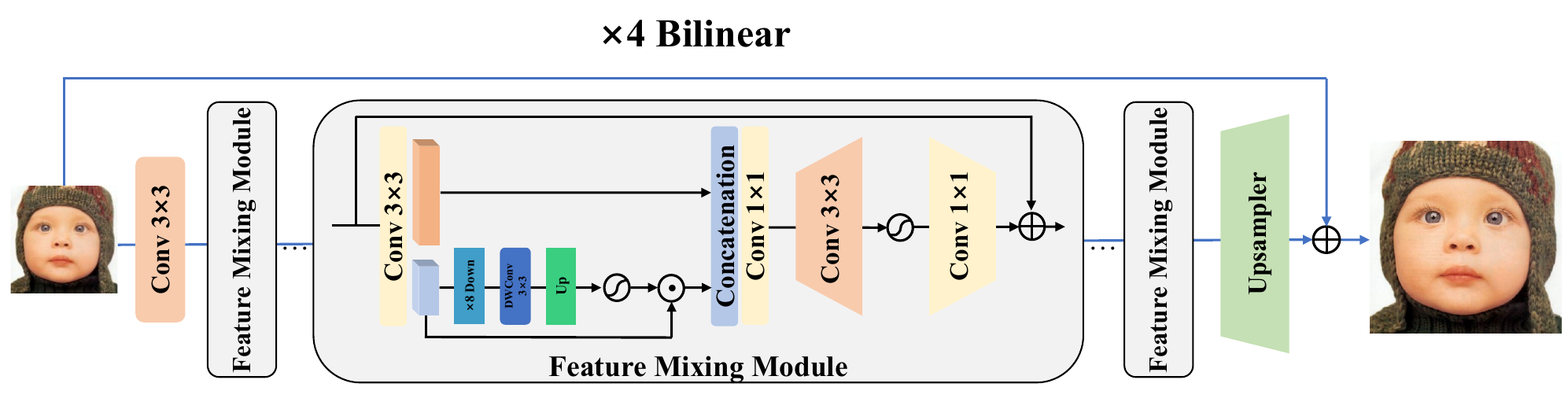}
    \caption{An overview of the proposed SAFMN++ model.}
    \label{fig:safmn}
\end{figure*}


\subsection{An efficient and fast network for super-resolution using convolution}
\label{sec:vpeg}

\begin{center}

\vspace{2mm}
\noindent\emph{\textbf{Teams VPEG}}
\vspace{2mm}

\noindent\emph{
Jiangtao Lv, Long Sun, Jinshan Pan, Jiangxin Dong,\\ Jinhui Tang
}

\vspace{2mm}

\noindent\emph{Nanjing University of Science and technology}

\end{center}


\paragraph{SAFMN++: Improved Feature Modulation Network for Real-Time Compressed Image Super-Resolution}

We introduce SAFMN++, an enhanced version of SAFMN~\cite{sun2023safmn} for solving real-time compressed image SR.
This solution is mainly concentrates on improving the effectiveness of the spatially-adaptive feature modulation (SAFM)~\cite{sun2023safmn} layer.
Different from the original SAFM, as shown in Fig~\ref{fig:safmn}, the improved SAFM (SAFM++) is able to extract both local and non-local features.
In SAFM++, a 3$\times$3 convolution is first utilized to extract local features and a single scale feature modulation is then applied to a portion of the extracted features for non-local feature interaction.

After this process, these two sets of features are aggregated by channel concatenation and fed into a 1$\times$1 convolution for feature fusion.

The proposed SAFMN++ is trained by minimizing a combination of the uncertainty-based MSE loss~\cite{uncentraintyloss,featup} and FFT-based L1 loss~\cite{MIMO} with Adam optimizer for a total of 500,000 iterations. 
We train the proposed SAFMN++ on the DIV2K~\cite{DIV2K} dataset. The cropped LR image size is $640\times640$ and the mini-batch size is set to 64. We set the initial learning rate to $3\times10^{-3}$ and the minimum one to $1\times10^{-7}$, which is updated by the Cosine Annealing scheme~\cite{cosine}.

Table~\ref{tab:efficiency} presents the efficiency study of SAFMN++.

\paragraph{A Simple Residual ConvNet with Structural Re-parameterization for Real-Time Super-Resolution}

The solution VPEG-R is shown in \cref{fig:vpegr}. The proposed method reduces the spatial resolution by a Pixel Unshuffle operation and uses a convolutional layer to transform the input LR image into the feature space, then performs performs feature extraction using 3 reparameterizable residual blocks (RepRBs), and finally reconstructs the final output by a PixelShuffle~\cite{espcn} convolution.

We use DIV2K~\cite{DIV2K} as the training data. In order to accelerate the IO speed during training, we crop the 2K resolution HR images to $640\times640$ sub-images, and the mini-batch size is set to 64.

\begin{table}[t]
    \centering
    \resizebox{\linewidth}{!}{
    \begin{tabular}{c c c c c}
    \toprule
    Method & Params [M] & FLOPs [G] & Runtime [ms] &Val. PSNR \\
    \midrule
    VPEG-S & 0.0662 &34.1587 & 11.5839 &23.29  \\
    VPEG-R & 0.0122 &1.556 & 2.2531 &22.77  \\
    \bottomrule
    \end{tabular}
    }
    \caption{SAFMN++: efficiency results. ``FLOPs” and ``Runtime” are tested on an LR image of size 540$\times$960 with an NVIDIA RTX3060.}
    \label{tab:efficiency}
\end{table}

\vspace{-2mm}

\paragraph{Implementation details}

We use PyTorch and a NVIDIA GeForce RTX 3090 GPU. The training process takes about 44 hours for SAFMN++, and two days for VPEG-R.

\begin{figure*}[t]
\centering
\captionsetup[subfigure]{position=bottom}
    \subcaptionbox{Training mode of the proposed network.\label{fig: proposed_network}}
    {\includegraphics[width=\textwidth]{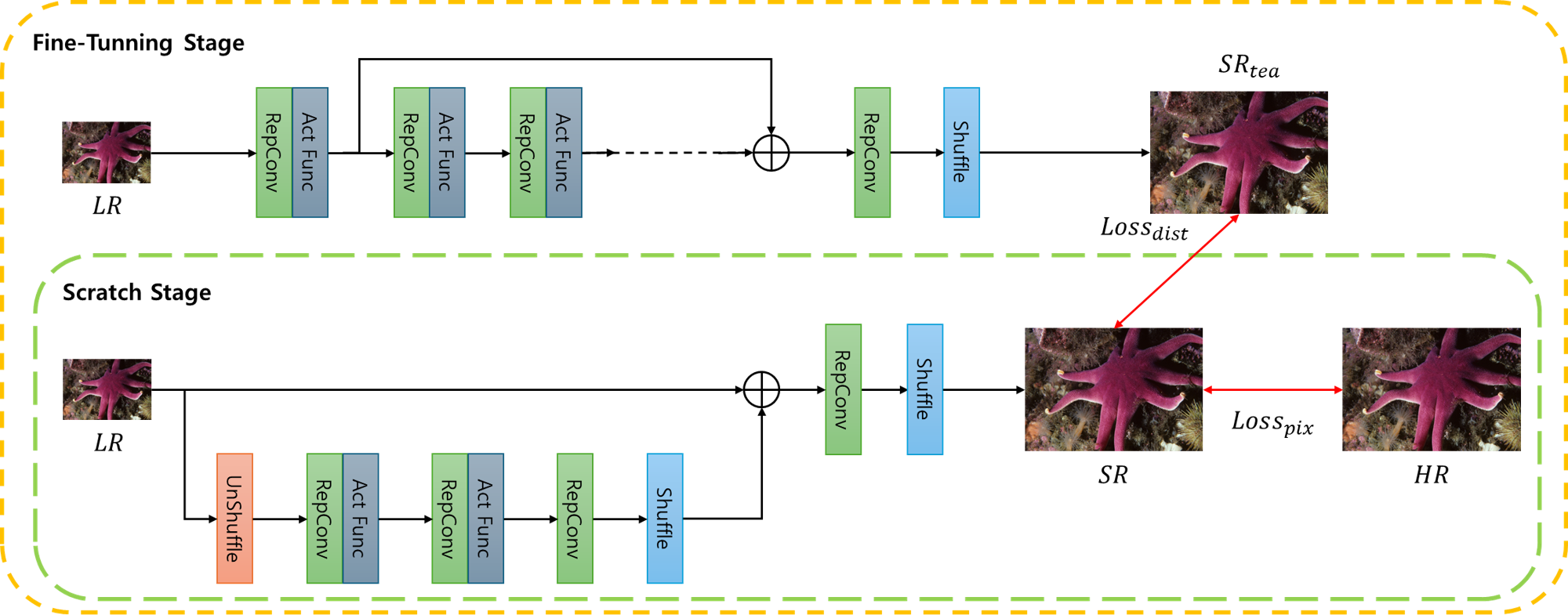}}
    \vfill
    \subcaptionbox{Inference mode of the proposed network. \label{fig: dist xcp test}}
    {\includegraphics[width=\textwidth]{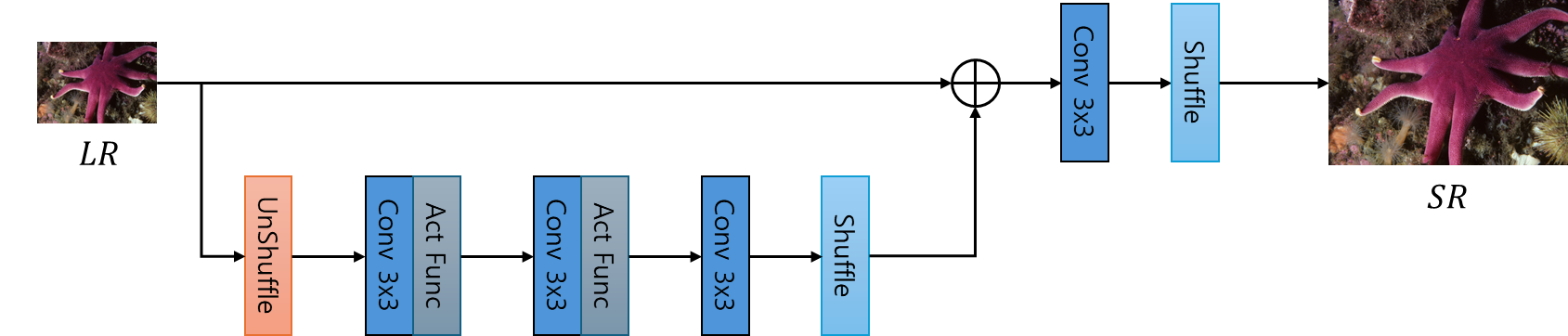}}
  \caption{CASR network structure proposed by team Z6.}
\end{figure*}


\subsection{CASR: Efficient Cascade Network Structure with Channel Aligned method for 4K Real-Time Single Image Super-Resolution}
\label{sec:CASR_z6}

\begin{center}

\vspace{2mm}
\noindent\emph{\textbf{Team Z6}}
\vspace{2mm}

\noindent\emph{
Kihwan Yoon$^1$,
Ganzorig Gankhuyag$^2$
}

\vspace{2mm}

\noindent\emph{
$^1$ The University of Seoul\\
$^2$ Korea Electronics Technology Institute (KETI)}

\end{center}



We initially reviewed the key factors essential for developing a network structure. Subsequently, we suggest a Cascade Upsampling network structure with Channel Alignment approach for image enhancement, which enhances performance and notably decreases processing time. Lastly, we designed an effective network and integrated reparameterization blocks and knowledge distillation methods to enhance performance without increasing the model's size~\cite{yoon2024casr}.  

We compared our proposed method with LRSRN~\cite{lightweight2023} which proposed work on the NTIRE 2023 real-time super-resolution challenge~\cite{conde2023efficient} in \cref{tab:results_z6}. The score value is calculated from the script of \cite{conde2023efficient}. Our proposed method overwhelms the previous method and we achieve 0.5678 ms inference time at RTX3090. 

\begin{table}[t]
    \centering
    \resizebox{\linewidth}{!}{
    \begin{tabular}{c c c c}
        \toprule
         Method \cite{lightweight2023}& Div2K Val & AIS2024 Val & Inference Time (ms) \\
         \midrule
         LRSRN \cite{lightweight2023} & 27.26 & 23.1 & 4.4890 \\
         Proposed method & 26.69 & 22.80 & 0.5678 \\
         \bottomrule
    \end{tabular}}
    \caption{Ablation study of team Z6.}
    \label{tab:results_z6}
\end{table}

\paragraph{Implementation details}

We used two different types of dataset: DIV2K and combined datasets.

\begin{itemize}
\item DIV2K: Well-known open dataset. DIV2K training data set used in the scratch training step.

\item Combined: The DIV2K training dataset is utilized during the initial training phase from scratch. In contrast, a composite dataset is used for the subsequent second stage. This combined dataset comprises the full DIV2K training set (800 images), the initial 1000 images from the Flickr training set, 121 samples from the GTA training sequences 00 to 19, the first 1000 images from the LSDIR dataset. To generate low resolution, we degrade the random cropped images with avif compression with various compression factors. For both training stages, we used random cropping, rotation 90, horizontal flip, and vertical flip augmentation.

\end{itemize}

We trained our model in three steps:

(1) Scratch train step: In the first step, our model was trained from scratch. The LR patches were cropped from LR images with eight mini-batch 98 x 98 sizes. Adam optimizer was used with a learning rate of 0.0005 during scratch training. The total number of epochs was set to 800. We use the $l1$ loss.

(ii) Second step: In the second step, the model was initialized with the weights trained in the first step. The distillation method used at this stage. The teacher model was trained with the combined dataset. The detailed illustrated example is shown in Fig. \ref{fig: dist xcp test}. Fine tuning with loss $l2$ improves the PSNR by 0.01 $\sim$ 0.02 dB. Also, we turn off the bias term of the reparametrization block at this stage. In this step, the initial learning rate was set to 0.00005 and the Adam optimizer was used along with a cosine warm-up. The total epoch was set at 800 epochs. 

(iii) Third step: In the third stage, the model was initialized using the weights trained in the previous step. In addition, the distillation technique was applied in this phase as well. The training hyper-parameters were kept identical to those in the second step. At this point, the bias term of the reparametrization block was deactivated, leading to a decrease in inference time by 0.2 ms. Although there was a slight reduction of 0.02 dB in the precision of the PSNR value, the overall score improved.

We refer the reader to the CASR~\cite{yoon2024casr} paper for more details.



\subsection{RVSR: Real-Time Super-Resolution with Re-parameterization and ViT architecture}
\label{sec:xjtu}

\begin{center}

\vspace{2mm}
\noindent\emph{\textbf{Team XJTU-AIR}}
\vspace{2mm}

\noindent\emph{
Zhiyuan Li, Hao Wei, Chenyang Ge
}

\vspace{2mm}

\noindent\emph{
Institute of Artificial Intelligence and Robotics, Xi'an Jiaotong University}

\end{center}

\begin{figure}[t]
    \centering
    \includegraphics[width=0.95\linewidth]{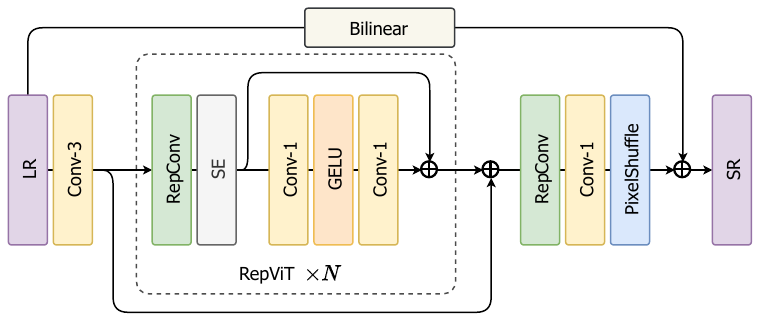} \\ 
    \vspace{0.1cm}
    \makebox[0.5\linewidth]{(a) Detailed architecture of RVSR by Team XJTU-AIR.}\\
    \includegraphics[width=0.95\linewidth]{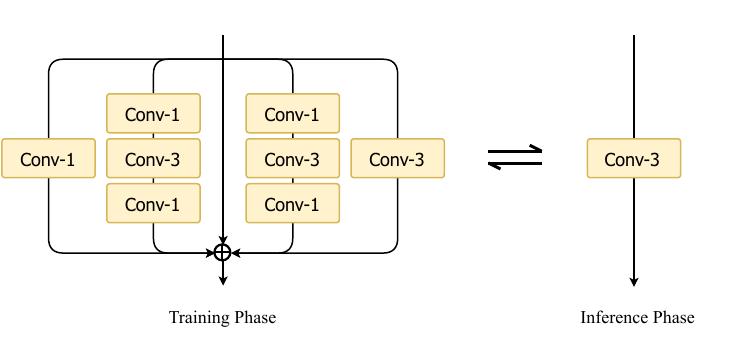} \\ 
    \vspace{0.1cm}
    \makebox[0.5\linewidth]{(b) The RepConv module} \\
    \caption{Overview of the proposed RVSR by Team XJTU-AIR.}
    \label{fig:xjtu}
\end{figure}

We propose a real-time image super-resolution method called RVSR, which is inspired by previous work \cite{Repvit, FMEN}. Our method leverages the efficient architectural designs of lightweight ViTs and the re-parameterization technique to achieve superior performance in real-time super-resolution tasks. 
RVSR first applies a 3$\times$3 convolution to convert the channel of feature map to the target size (16). Then, RVSR employs 8 stacked RepViT \cite{Repvit} blocks to perform deep feature extraction. As shown in Fig. \ref{fig:xjtu}~(a), the RepViT blocks integrate the efficient architectural designs of lightweight ViTs. Inspired by \cite{FMEN}, RVSR employs the RepConv module to improve the SR performance while maintaining low complexity, as shown in Fig. \ref{fig:xjtu}~(b). 

We conducted an end-to-end training of the RVSR model for 5000 epochs, employing a batch size of 32 and optimizing by minimizing the MSE loss with the Adam optimizer. For inference, we re-parameterized the model using standard 3x3 convolutions, as illustrated in Fig. \ref{fig:xjtu}~(b).

\vspace{-2mm}

\paragraph{Implementation details}

The method is implemented in PyTorch. For optimization, we utilize the Adam optimizer with $\beta_1=0.99$ and $\beta_2=0.999$. The learning rate is set to $5 \times 10^{-4}$ for the first 1000 epochs, after which it linearly decays until reaching $1 \times 10^{-6}$. 

We trained RVSR on DIV2K dataset (800 images), Flickr2K dataset (2650 images) and LSDIR dataset (first 1000 images). For generating low-resolution images, we employed Lanczos downsampling and AVIF compression, with compression factors ranging from QP 31 to 63. During training, we used random cropping, rotations, and flips augmentations. Besides, the images are normalized to the range [-1, 1]. 
The experiments were conducted on a Nvidia GeForce RTX 3090 GPU, with the input size set to 960$\times$540. MACs: 15.62 (G), 1883 MACs per pixel, runtime: 12.54 ms (FP32) and 7.36 ms (FP16).

\begin{figure*}[t]
\centering
\includegraphics[width=0.75\textwidth]{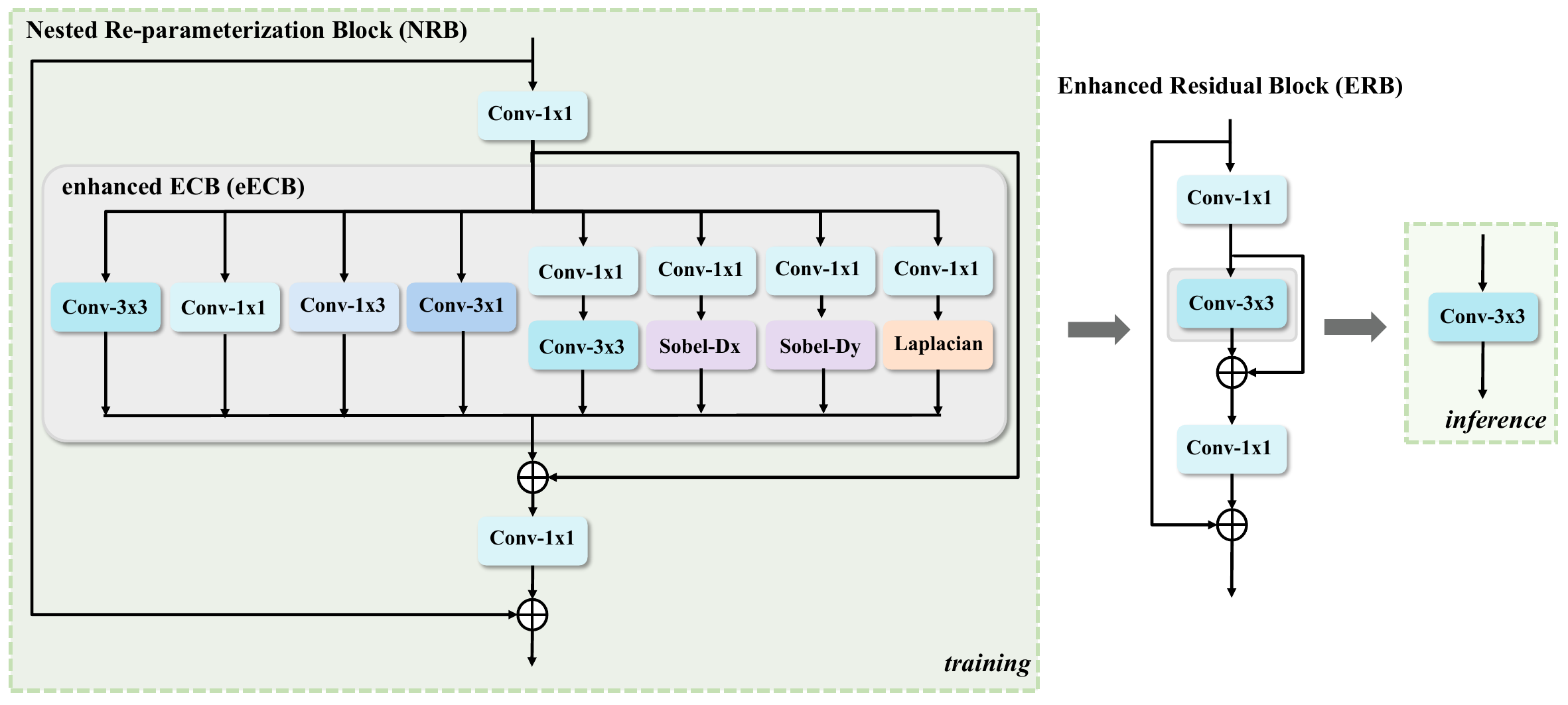}
\caption{Detail network of the proposed Nested Re-parameterization Block (NRB), Team LeRTSR.}
\label{fig:block}
\vspace{-0.2cm}
\end{figure*}


\subsection{Anchor-based Nested UnshuffleNet for Real-
time Super-Resolution (ANUNet)}
\label{sec:lenovo}

\begin{center}

\vspace{2mm}
\noindent\emph{\textbf{Team LeRTSR}}
\vspace{2mm}

\noindent\emph{
Menghan Zhou, Yiqiang Yan
}

\vspace{2mm}

\noindent\emph{
Lenovo Research}

\end{center}

\begin{figure}[t]
\centering
\includegraphics[width=1\columnwidth]{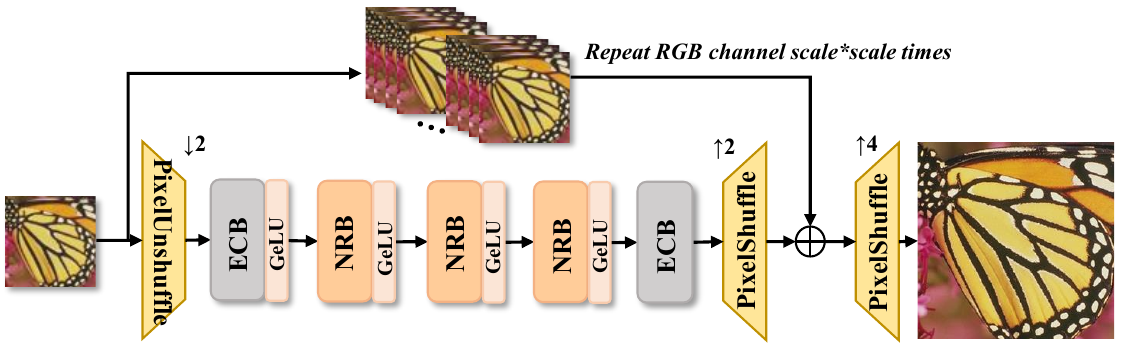}
\caption{The framework of the proposed Anchor-based Nested UnshuffleNet for Real-time Super-Resolution (ANUNet).}
\label{fig:pipeline}
\vspace{-0.2cm}
\end{figure}

We propose Anchor-based Nested UnshuffleNet for Real-time Super-Resolution (ANUNet). As shown in \cref{fig:pipeline}, the pixel-unshuffle technique \cite{pixelunshuffle} is used to reduce the resolution of the image and increase the channel
dimension. This design allows for a reduction in the computational overhead of the network while preserving the constant volume of information. After an ECB \cite{ECBSR} + GeLU module, the main module is composed of a sequence of Nested Re-parameterization Block (NRB) + GeLU activation, which serves to extract and refine features in a progressive manner. Then, an ECB layer is adopted to transfer features, followed by an upsampling layer for recovering the resolution to LR. While an anchor-based residual learning is applied to directly repeat the RGB channels 16 times in LR space to generate anchors. Finally, a pixel shuffle layer is is used to reconstruct the final HR output.

Different from \cite{FMEN} and \cite{ECBSR}, we design a nested structure, named  Nested Re-parameterization Block (NRB). \cref{fig:block} illustrates the proposed NRB. 
In the training stage, the NRB employs a nested structure, the outer structure is the ERB RepBlock in the Enhanced Residual Block (ERB) first proposed by \cite{FMEN}, the inner structure is an enhanced Edge-oriented Convolution Block (eECB), which includes multiple branches, and can be merged into one normal convolution layer in the inference stage. Performance remains unaffected after re-parameterization in this design.


\paragraph{Implementation details}

We use DIV2K \cite{DIV2K} and Flickr2K for training. To generate the compressed LR images, we use AVIF to process the above datasets with the random QP ranges between 31 and 63. Besides, standard augmentations that include all variations of flipping and rotations are also used to improve performance. Additionally, the number of feature channels is set to 28, and the scale of pixel unshuffle and pixel shuffle in the sub-branch is set to 2. After the training, we re-parameterize the model into a network structure (ECB and NRB modules) with regular 3x3 convolutions.

The model is conducted using the PyTorch framework with one NVIDIA A100 40G GPU. Specifically, the training is divided into three stages:

1. Initially, the model is trained from scratch with 480×480 patches randomly cropped from high resolution (HR) images with a mini-batch size of 64. We apply a combination of Charbonnier loss \cite{Charloss} and FFT-based frequency loss \cite{MIMO} function  for reconstruction. The network is trained for 1000k iterations using the Adam optimizer, with a learning rate 1 × $10^{-3}$ decreasing to 1 × $10^{-6}$ through the cosine scheduler.

2. In the second stage, the model is initialized with the pre-trained weights from the first stage on the same training data as stage 1.  Inspired by \cite{swin2sr}, the auxiliary loss and high-frequency loss are added to our training. Instead of the downsampling bicubic operator used in \cite{swin2sr}, Lanczos is applied to maintain consistency with the downsampling method in AVIF. The network parameters are optimized for 1000k iterations  with the MultiStepLR scheduler, where the initial learning rate is set to 5 × $10^4$ and halved at {200k, 400k, 800k}-iteration. 

3. The model is fine-tuned using the L2 loss and FFT loss. The other settings are the same as in stage 2. The network is trained for 1000k iterations in this stage.



\begin{figure*}[t]
\centering  
\begin{tabular}{c}
     \includegraphics[width=13cm]{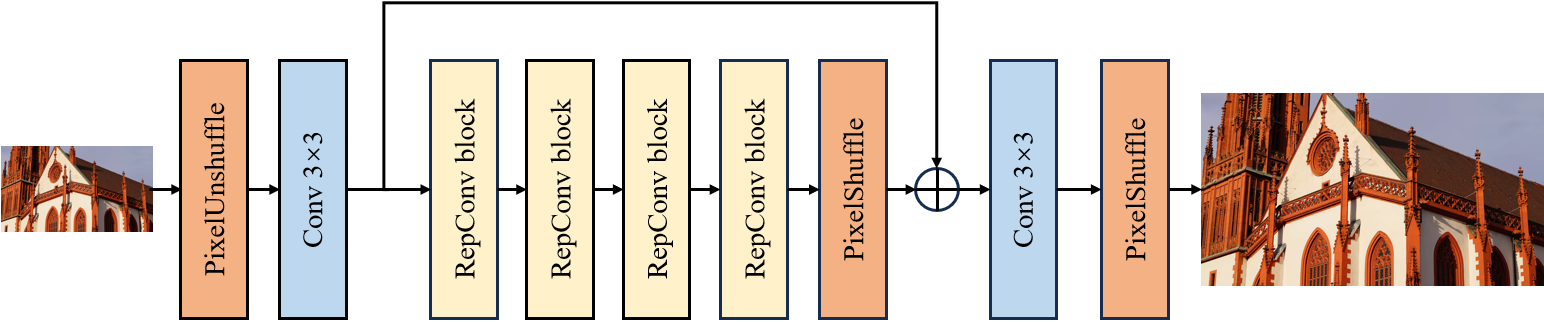} \\
     The training mode of proposed network \\
     \includegraphics[width=13cm]{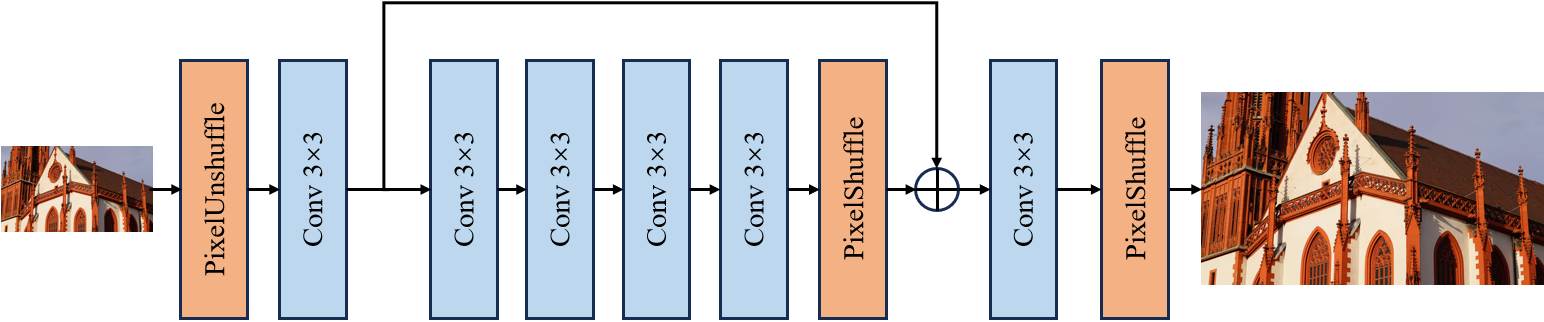} \\
     The inference mode of proposed network \\
\end{tabular}
\label{fig:FZUQXT_diagram}
\caption{Illustration of proposed RESR (Team FZUQXT).}
\end{figure*}


\subsection{RESR: Reparameterized and Edge-oriented Network for Real- Time Image Super-Resolution}
\label{sec:FZUQXT}

\begin{center}

\vspace{2mm}
\noindent\emph{\textbf{Team FZUQXT}}
\vspace{2mm}

\noindent\emph{
Xintao Qiu~$^1$,
Yuanbo Zhou~$^1$,
Kongxian Wu~$^1$,
Xinwei Dai~$^1$,
Hui Tang~$^1$,
Wei Deng~$^2$,\\
Qingquan Gao~$^1$,
Tong Tong~$^1$
}

\vspace{2mm}

\noindent\emph{
$^1$ Fuzhou University\\
$^2$ Imperial Vision Technology}

\end{center}



We propose a real-time image super-resolution based on re-parameterization and edge extraction. We use pixel unshuffle to reduce the image resolution and increase the channel dimension. This design reduces the computational cost of the network while keeping the amount of information constant. Meanwhile, we propose a reparameterized image edge extraction block that extracts features in parallel through multiple paths in the training phase, including 3×3 and 1×1 convolution for channel expansion and compression, as well as sobel and laplacian filters for acquiring information about image edges and textures. 

In the inference stage, multiple operations can be combined into a 3×3 convolution. The performance of 3×3 convolution is improved without introducing any extra cost.


\begin{figure}[t]
\centering  
\includegraphics[width=\linewidth]{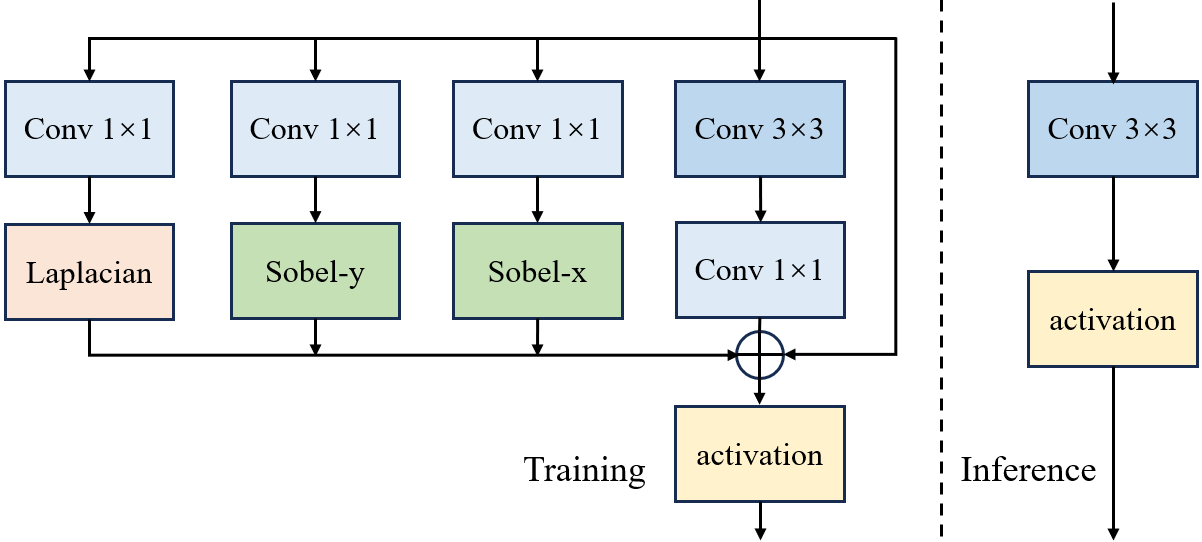}
\caption{The architecture of RepConv block (Team FZUQXT).}
\label{2}
\end{figure}

\paragraph{Efficiency metrics} Considering the challenge input image, the model has 7.0171 GMACs, 14.0341 GFLOPs and the runtime is 1.64ms (using FP16).

\paragraph{Implementation details}

The datasets we used include the DIV2K training set (800
images) and the Flicker2K training set (2650 images). To
increase the speed of IO, we split the original HR (high
resolution) and LR (low resolution) images into multiple
corresponding 600×600 and 150×150 patches. We randomly flipped these patches by flipping them horizontally, vertically and rotating them by 90-degrees to augment the data. 

We use PyTorch and a RTX 3090 GPU (24GB). The models are optimized using Adam with Cosine Warmup. The total duration of the training process is $\approx48$hrs.

In the first training stage, we train our model from scratch. The LR patches cropped from LR images with 128x128 image size and 64 mini-batch. The Adam optimizer uses a 0.0005 learning rate. The cosine warm-up scheduler sets a 0.1 percentage warmup ratio. The total number of epochs in this stage is set to 800. 

In the second stage, we initialize the model with the weights trained in the previous stage. In this step, the initial learning rate is set as 0.0001. The cosine warm-up scheduler is set with a 0.1 percentage warm-up ratio. The total number of epochs is set to 200 epochs.



\subsection{Real Time Swift Parameter-free Attention Network for 4x Image Super-Resolution}
\label{sec:xiaomi}

\begin{center}

\vspace{2mm}
\noindent\emph{\textbf{Teams XiaomiMM C3}}
\vspace{2mm}

\noindent\emph{
Bingnan Han,
Hongyuan Yu,
Zhuoyuan Wu,
Cheng Wan,
Yuqing Liu,
Haodong Yu,
Jizhe Li, 
Zhijuan Huang, 
Yuan Huang, 
Yajun Zou, 
Xianyu Guan,
Qi Jia,
Heng Zhang, 
Xuanwu Yin, 
Kunlong Zuo
}

\vspace{2mm}

\noindent\emph{
$^1$ Multimedia Department, Xiaomi Inc.\\
$^2$ Georgia Institute of Technology\\
$^3$ Dalian university of technology}

\end{center}


\paragraph{Real Time Swift Parameter-free Attention Network for 4x Image Super-Resolution}

We propose a convolutional neural network combining swift parameter-free attention block (SPAB) for image SR, the suggested model has very few parameters and fast processing speed for 4x image super resolution. 

As shown in \cref{fig:SPANR}, SPAN consists of 2 consecutive SPABs and each SPAB block extracts progressively higher-level features sequentially through three convolutional layers with $C'$-channeled $H' \times W'$-sized kernels (In our model, we choose $H' = W' = 3$.). The extracted features $H_i$ are then added with a residual connection from the input of SPAB, forming the pre-attention feature map $U_i$ for that block. The features extracted by the convolutional layers are passed through an activation function $\sigma_a(\cdot)$ that is symmetric about the origin to obtain the attention map $V_i$. The feature map and attention map are element-wise multiplied to produce the final output $O_i=U_i \odot V_i$ of the SPAB block, where $\odot$ denotes element-wise multiplication. We use $W_i^{(j)}\in R^{C' \times H'\times W'}$ to represent the kernel of the $j$-th convolutional layer of the $i$-th SPAB block and $\sigma$ to represent the activation function following the convolutional layer.  Then the SPAB block can be expressed as:

\begin{equation}
    \begin{aligned}
        O_i&=F_{W_i}^{(i)}(O_{i-1})=U_i \odot V_i,\\
        U_i&=O_{i-1}\oplus H_i, \quad V_i=\sigma_a(H_i),\\ 
        H_i&=F_{c,W_i}^{(i)}(O_{i-1}),\\
        &=W_i^{(3)}\otimes\sigma(W_i^{(2)}\otimes\sigma(W_i^{(1)}\otimes O_{i-1})),
    \end{aligned}
    \label{equ:SPAB}
\end{equation}

where $\oplus$ and $\otimes$ represent the element-wise sum between extracted features and residual connections,  and the convolution operation, respectively.  $F_{W_i}^{(i)}$ and $F_{c,W_i}^{(i)}$ are the function representing the $i$-th SPAB and the function representing the $3$ convolution layers of $i$-th SPAB with parameters   $W_i=(W_i^{(1)},W_i^{(2)},W_i^{(3)})$, respectively. $O_0=\sigma(W_0\otimes I_{\text{LR}})$ is a $C'$-channeled $H \times W$ feature map from the $C$-channeled $H \times W$-sized low-resolution input image $I_{LR}$ undergone a convolutional layer with $3 \times 3$ sized kernel $W_0$.  This convolutional layer ensures that each SPAB has the same number of channels as input.  The whole SPAN neural network can be described as

\begin{equation}
    \begin{aligned}
        I_{\text{HR}}&=F(I_{\text{LR}})=\text{PixelShuffle}[W_{f2}\otimes O)],\\
        O&=\text{Concat}(O_0,O_1,O_5,W_{f1}\otimes O_6),
    \end{aligned}
    \label{equ:SPAN}
\end{equation}

where $O$ is a $4C'$-channeled $H \times W$-sized feature map with multiple hierarchical features obtaining by concatenating $O_0$ with the outputs of the first, fifth, and the convolved output of the sixth SPAB blocks by $C'$-channeled $3 \times 3$-sized kernel $W_{f1}$.  $O$ is processed through a $3\times 3$ convolutional layer to create an $r^2C$ channel feature map of size $H\times W$. Then, this feature map goes through a pixel shuffle module to generate a high-resolution image of $C$ channels and dimensions $rH\times rW$, where $r$ represents the super-resolution factor. The idea of computing attention maps directly without parameters from feature extracted by convolutional layers, led to two design considerations for our neural network: the choice of activation function for computing the attention map and the use of residual connections, more details about activation function and SPAB module are in~\cite{wan2023swift}.


\begin{figure}[t]
    \centering
    \includegraphics[width=\linewidth]{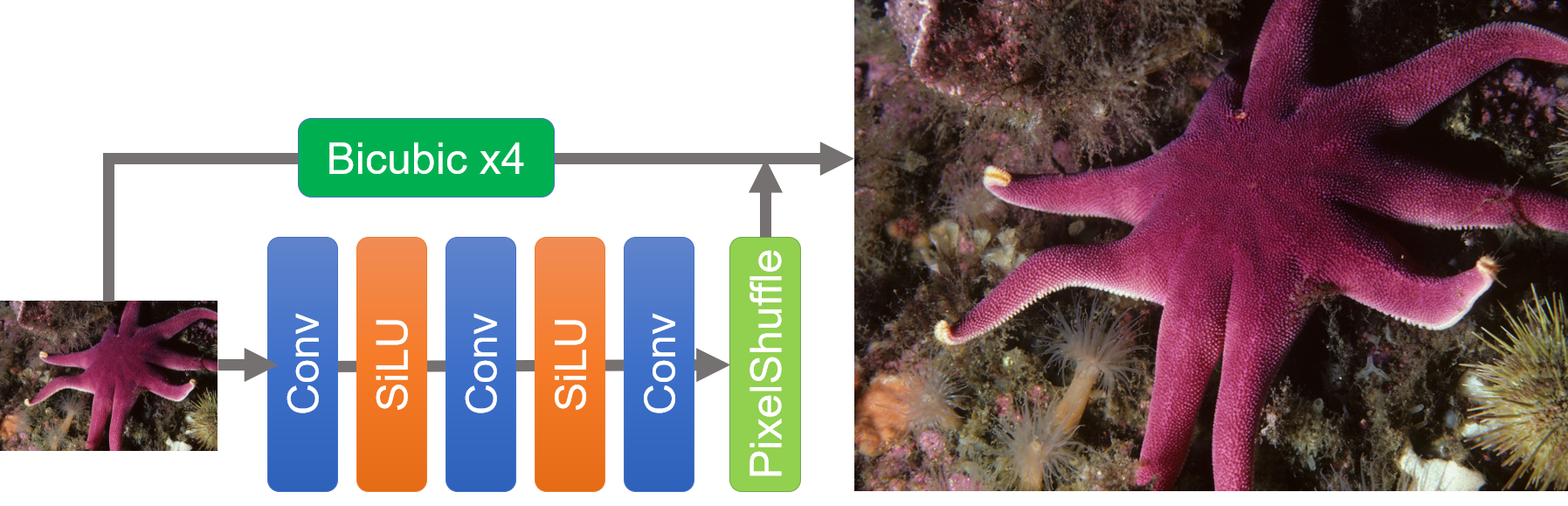}
    \caption{Network architecture of C3.}
    \label{fig:c3}
\end{figure}

\begin{figure*}[t]
    \centering
    \includegraphics[width=0.75\linewidth]{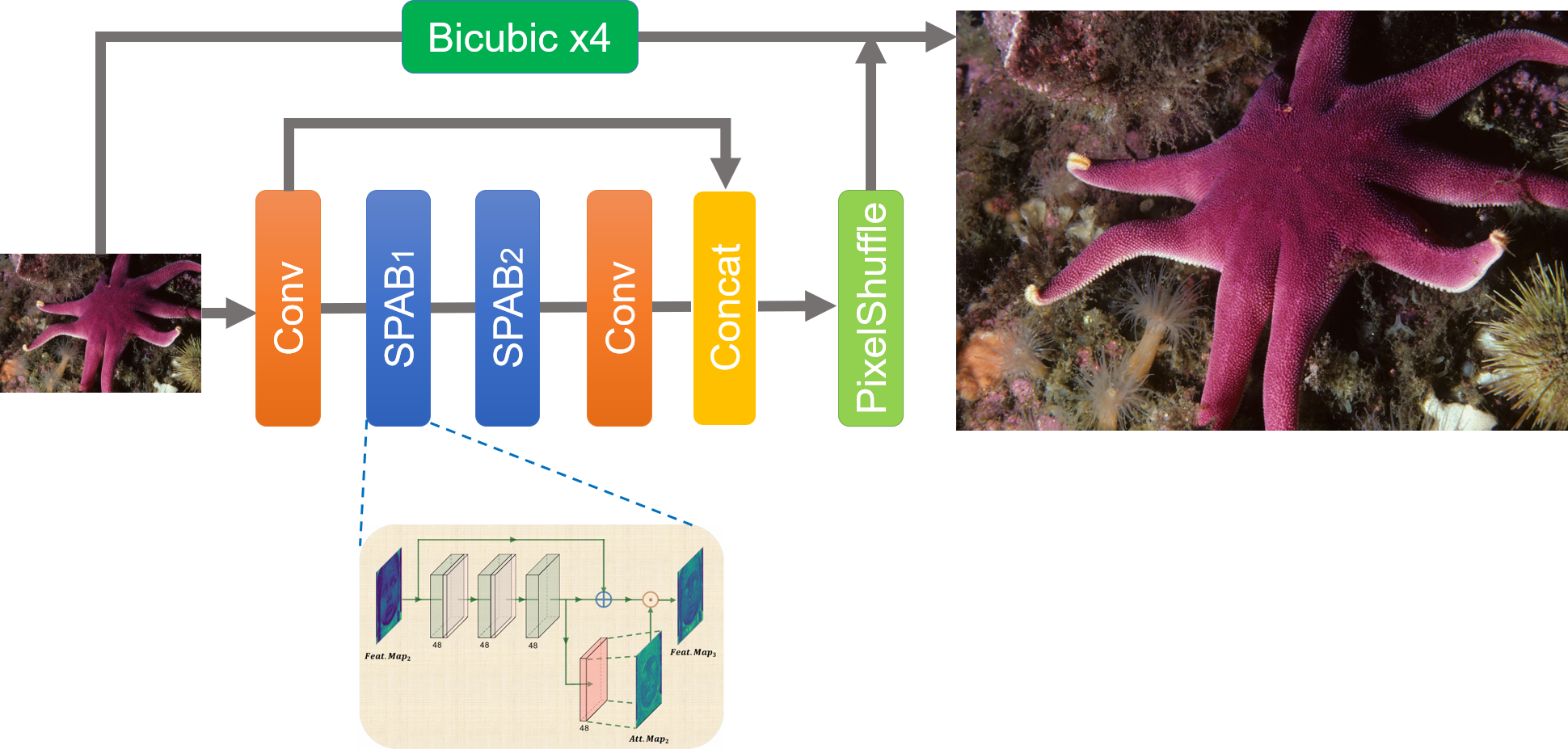}
    \caption{Network architecture of SPANR proposed by Team XiaomiMM.}
    \label{fig:SPANR}
\end{figure*}

\paragraph{C3 network for 4x image super-resolution}

A three-layer convolutional neural network for image SR, the suggested model has very few parameters and fast processing speed for 4x image super resolution. This model has 12.39 GFLOPs and 0.024 M parameters. The model is shown in \cref{fig:c3}.

\paragraph{Implementation details}

Both models use HAT-L\cite{chen2023activating} 4x pre-trained network for knowledge distillation.

\begin{itemize}
    \item \textbf{Framework:} Pytorch
    \item \textbf{Optimizer and Learning Rate:} We implement the network with PyTorch (BasicSR framework). The optimizer is Adam with learning rate as $10^{-4}$. 
    \item \textbf{GPU:} RTX A100
    \item \textbf{Datasets:} We randomly collect the videos from the Internet, and randomly compress them with different QP.
    \item \textbf{Training Time:} We initially utilize L1 loss along with Grad loss for the first step training with 500000 iterations, then for the second step training, we use MSE loss combined with Grad loss with 250000 iterations.
\end{itemize}

\begin{figure*}[t]
    \centering
    \includegraphics[width=0.9\linewidth]{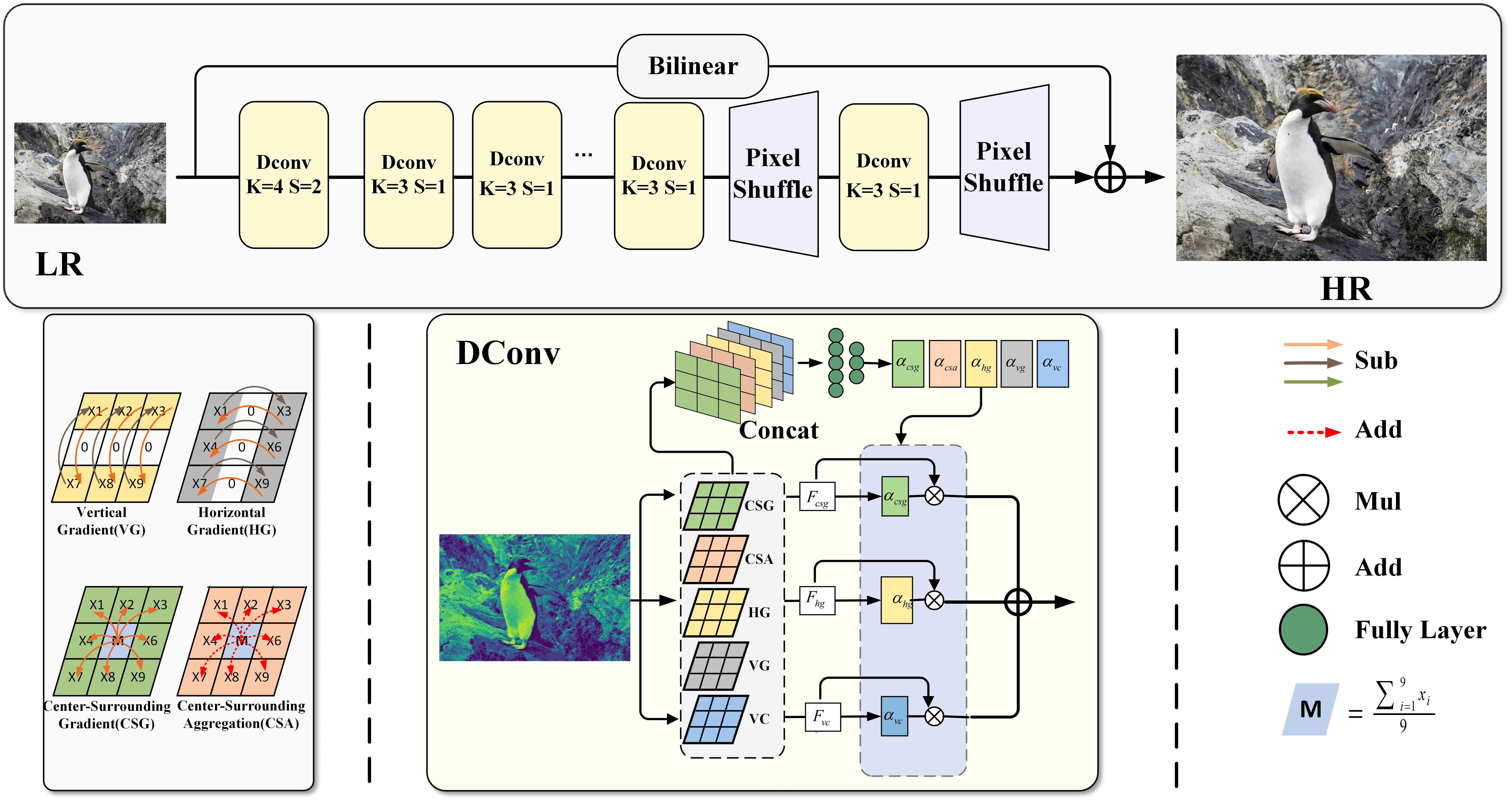}
    \caption{Diagram of the framework proposed by Team USTC Noah. The method utilizes a single DecoupleConv (DConv) with a kernel size of 4 and a stride of 2 to form the feature mapping layer. Concurrently, we construct the feature learning layer using three DConvs, each with a kernel size of 3 and a stride of 1. The resolution of the features is altered through the application of Pixel shuffle.}
    \label{fig:noah}
\end{figure*}


\subsection{Efficient Real-Time Image Super-Resolution Via Decouple Convolution}
\label{sec:noah}

\begin{center}

\vspace{2mm}
\noindent\emph{\textbf{Team USTC Noah Terminal Vision}}
\vspace{2mm}

\noindent\emph{
Long Peng~$^1$, Jiaming Guo~$^2$, Xin Di~$^1$, Bohao Liao~$^1$, Zhibo Du~$^1$, Peize Xia~$^1$, Renjing Pei~$^2$, Yang Wang~$^1$, Yang Cao~$^1$, Zhengjun Zha~$^1$
}

\vspace{2mm}

\noindent\emph{
$^1$ University of Science and Technology of China \\
$^2$ Huawei Noah’s Ark Lab}

\end{center}


To enhance the network’s perception of gradients and
contrast, we have refined the existing vanilla convolution
unit by performing feature decoupling within local regions. We innovatively introduce gradient (sub) operators
and aggregation (add) operators to convolution to capture
detail and contrast relevant properties. Specifically, we have introduced differential operations into the convolutional process to preemptively capture horizontal, vertical, and central-surrounding directions. Furthermore, we have incorporated an aggregation (add) operation into the convolution to boost the network’s sensitivity to statistical features. The method is shown in \cref{fig:noah}.

We initially applied the DecoupleConv (with kernel=4
and stride=2) to reduce the spatial resolution while si-
multaneously increasing the number of channels. Subse-
quently, we employed four decoupled convolutions with
reparameterization, which we designed for feature learn-
ing. We then utilized pixel shuffle on the features to up-
scale the image resolution to its original low resolution
(LR) size. Following this, a single decoupled convolu-
tion with reparameterization was used for feature map-
ping. Finally, another pixel shuffle operation was applied
to achieve a 4x super-resolution result.


\paragraph{Implementation details}

We utilized solely the DIV2K dataset and applied the
official compression methods to compress the images at
various levels, specifically at 31, 39, 47, 55, and 63 compression levels, amounting to a total of five different degrees of compression.

\textbf{Training:} We utilized the Adam optimizer with an initial learning rate of 5e-4, performing a total of 1e7 iterations. We employed the stepDecayLR learning rate strategy, which involves a decay every 2e6 iterations with a decay factor of 2. On each card, we set the batch size to 32, resulting in a cumulative batch size of 32*8 across all cards. The training was conducted over approximately 7 days, distributed across 8 V100 GPUs.

\textbf{Inference:} Prior to inference, it is necessary to perform an equivalent transformation of the parameters.



\subsection{A lightweight Super-resolution Algorithm Based on Re-Parameterization}
\label{sec:basicvision}

\begin{center}

\vspace{2mm}
\noindent\emph{\textbf{Teams BasicVision, CMVG, IVP}}
\vspace{2mm}

\noindent\emph{
Min Yan~$^1$,
Xin Liu~$^1$,
Qian Wang~$^1$,
Xiaoqian Ye~$^1$,
Zhan Du~$^1$,
Tiansen Zhang~$^2$
}

\vspace{2mm}

\noindent\emph{
$^1$ China Mobile Research Institute\\
$^2$ Min Zu University of China}

\end{center}


\paragraph{A lightweight Super-resolution Algorithm Based on Re-Parameterization}

We propose an efficient super-resolution network, which contains four convolutions and an unshuffle block. First, the network uses a convolutional operation for feature extraction. Then, it utilizes two re-parameterization modules to extract edge and detailed information. The re-parameterization module increases the number of parameters during training, but it is replaced by a single convolution to reduce computational complexity and memory usage during testing. The re-parameterization module we use can extract more edge and detailed information. Subsequently, another convolution operation is used to increase the number of channels to 48, which facilitates the subsequent four-fold super-resolution. 

\begin{figure}[t]
\begin{center} 
\includegraphics[width=\linewidth]{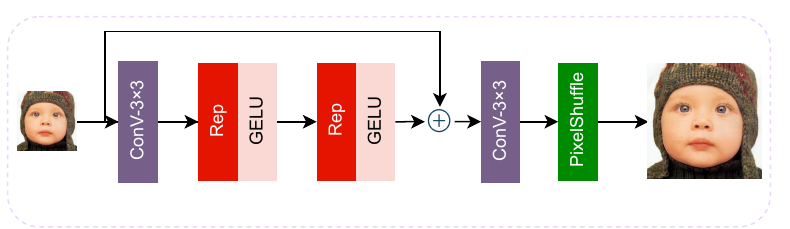} 
\caption{Overall framework of BasicVision.} 
\label{fig:basicvisionfig1}
\end{center}
\vspace{-5mm}
\end{figure}

\begin{figure}[t]
\begin{center} 
\includegraphics[width=\linewidth]{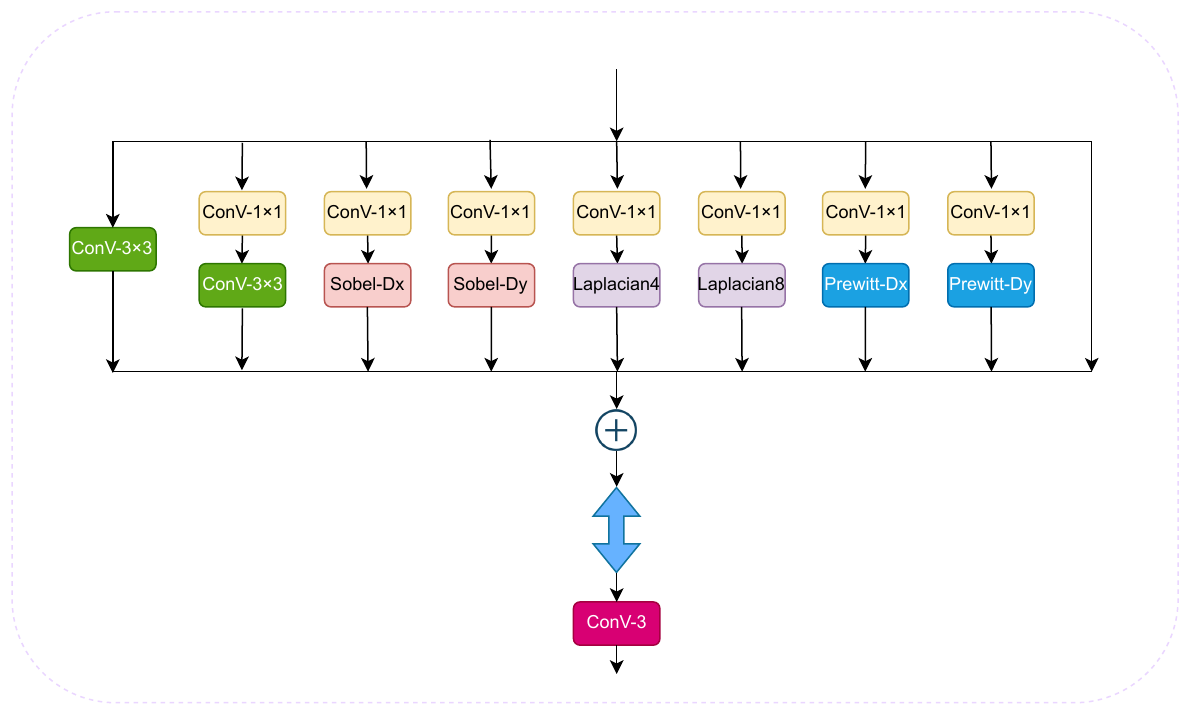} 
\caption{The re-parameterization module of BasicVision.} 
\label{fig:basicvisionfig2}
\end{center}
\vspace{-5mm}
\end{figure}

\begin{figure}[t]
\begin{center} 
\includegraphics[width=\linewidth]{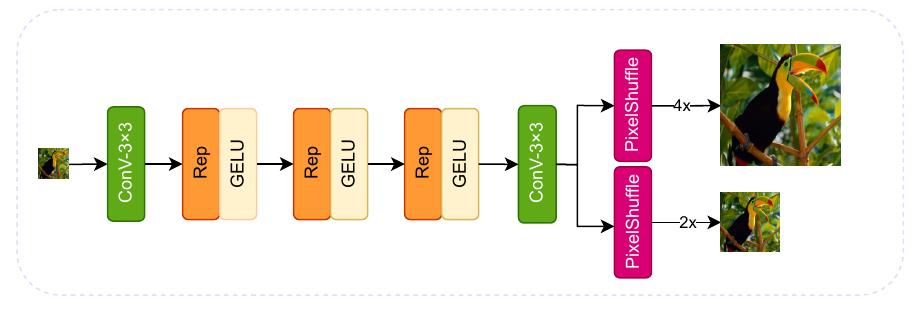} 
\caption{The framework of RDEN (IVP).} 
\label{fig1}
\end{center}
\vspace{-5mm}
\end{figure}

\begin{figure*}[t]
  \centering
  \includegraphics[width=0.9\textwidth]{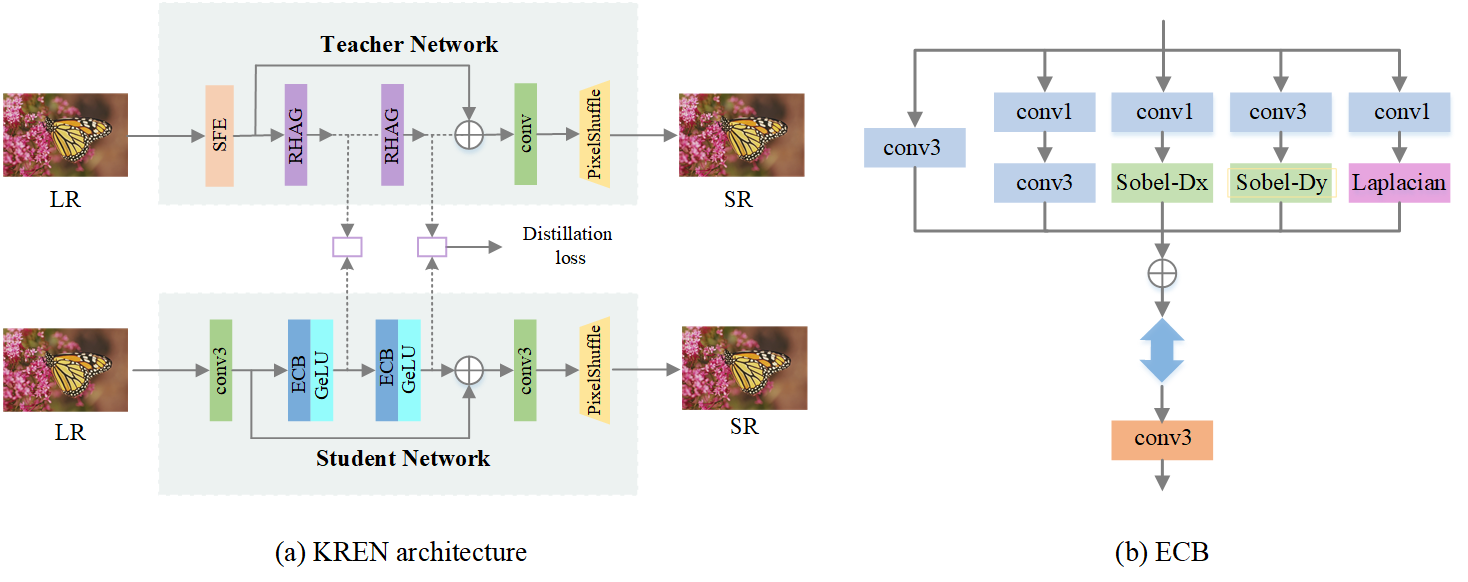}
  \caption{Overall architecture of KREN proposed by CMVG.}
  \label{fig:1}
\end{figure*}

Finally, we use an unshuffle block to make the channel-to-space transition. The whole network is shown in Figure~\ref{fig:basicvisionfig1}, and the convolutional layers in the middle(red) are two re-parameterization modules. The re-parameterization module we used is shown in Figure~\ref{fig:basicvisionfig2}. 

The local frequency loss (FFL) based on Fast Fourier Transform (FFT) \cite{focal} enables the model to dynamically prioritize challenging frequency components while diminishing the influence of easily synthesizable ones. This optimization objective supplements current spatial losses and effectively guards against the degradation of crucial frequency details caused by inherent biases in neural networks. We use the following FFT loss in our training:

\begin{equation}
    L_{FFT} = \|FFT(X_S^{SR})-FFT(X_S^{HR})\|
  \label{eq:3}
\end{equation}

Drawing inspiration from SPSR \cite{ma2020structure}, we propose a gradient loss that aids the model in accurately evaluating the local sharpness intensity of images. We use gradient loss in our training,  which is represented as follows:

\begin{equation}
    L_{GM} = \|GM(X_S^{SR})-GM(X_S^{HR})\|
  \label{eq:4}
\end{equation}

The overall loss for training the network is defined as:

\begin{align}
  L_S &= \alpha\|X_{S}^{SR4}-X_{S}^{HR4}\|+\gamma L_{\text{GM}}+\delta L_{\text{FFT}}
\label{eq:5}
\end{align}

\paragraph{Real-Time Super-Resolution with auxiliary loss}

The network we used is shown in Figure~\ref{fig1}, which contains three reparameterized modules, and an auxiliary head with upscale factor 2. We use the ECB \cite{zhang2021edge} model as the reparameterized module which can achieve competitive performance without computation overhead. Alongside the 4x super-resolution task, we introduce a 2x upsampling head for the 2x SR task. This additional task offers multiple benefits: it functions as a form of simulated annealing, allowing for potential escape from local minima; it serves as a prior, enhancing the delineation of our primary task. The loss associated with the 2x supervision is represented as follows: 
\begin{equation}
\label{eq:Positional Encoding}
\begin{split}
    L_{X2} = \|X_{S}^{SR2}-X_{S}^{HR2}\|
\end{split}
\end{equation}
where the {$X_{S}^{SRx2}$} denotes the output from 2x upsampling head, and {$X_{S}^{HRx2}$} are corresponding 2x HR image. Note that we cut the 2x super resolution model off in the testing phase, and the network consists of only five convolutions and one 4x upsampling head.

\vspace{-2mm}

\paragraph{Jointly supervision knowledge distillation network for efficient super-resolution}

We propose a efficient super-resolution network named KREN based on knowledge distillation and re-parameterization, as shown in Figure~\ref{fig:1}. 

The KREN model is composed of a teacher network and student network, we use the superior SR model HAT \cite{chen2023activating} as teacher network. The distillation training provides additional effective supervision information for student training, and enhances the performance and generalization ability of student network. The student network is composed of two convolution layers and two re-parameterization \cite{ECBSR} blocks ECB. The ECB block with complex structure is used in training phase, while it can be merged into a 3*3 convolution layer for speeding up inference speed during the inference phase. The re-parameterization strategy can effectively improve the feature diversity and boost the feature extraction ability of SR model. In addition, we propose a jointly supervision loss that consists the focal frequency loss(FFL) \cite{fft}, gradient map loss (GM) \cite{GM} , distillation loss and L1 loss. We extract features from the 1st and 3rd blocks of the teacher model, and features from each ECB block to calculate the distillation loss. The constraints on gradients and frequency domain helps super-resolved high quality images.We also propose a multi-stage progressive training strategy to gradually improves the reconstruction quality. The number of feature maps in student network is set to 14.

\begin{table}[t]
    \centering
    \renewcommand{\thetable}{1}
    \resizebox{\linewidth}{!}{
    \begin{tabular}{lcccccc}
    \toprule
        \text{Methods} & \text{Time [ms]} & \text{Params [M]} & \text{FLOPs [G]} & \text{Acts [M]} & \text{GPU Mem [M]} \\ 
        \midrule
        IMDN \cite{hui2019imdn} & 23.508 & 0.894 & 58.430 & 154.141 & 707.767 \\ 
        RFDN \cite{liu2020residual} & 18.569 & 0.433 & 27.046 & 112.034 & 791.928 \\ 
        RLFN \cite{kong2022residual} & 12.019 & 0.317 & 19.674 & 80.045 & 470.753 \\ 
        DIP \cite{yu2023dipnet}   & 10.049 & 0.243 & 14.886 & 72.9672 & 497.287 \\ 
    \bottomrule
    \end{tabular}
    }
    \caption{The lightweight metrics study by Teams BasicVision, CMVG, IVP. The "Time" denotes the average inference time. The ``Params" is the total number of parameters. The ``FLOPs" and ``Acts" are calculated on 256x256 images. The ``GPU Mem" represents the GPU memory during the inference.}
    \label{tab:1}
\end{table}

\vspace{-2mm}

\paragraph{Implementation details}
We train our model on DIV2K\cite{DIV2K}, Flickr2K\cite{Flickr2K} and GTA\cite{GTA} datasets, and utilize multi-stage training based on Pytorch on NVIDIA V100. The patch size in each training stage is selected from [256,384,512,640]. The mini-batch size is set to 64, and MSE, GM loss\cite{ma2020structure}, and FFT loss\cite{focal} are used as target loss functions. Each stage except for the first stage is fine-tuned based on the result of the previous stage, training for 500 epochs utilizing the Adam algorithm, beginning with a learning rate of $5 \times 10^{-4}$ and gradually decreasing to $5 \times 10^{-5} $following the cosine scheduler.

For the distillation approach (KREN) the training details are described as follows: \\
\noindent \textbf{Stage1.} Training teacher network.The teacher network is trained from scratch with teacher loss.\\
\textbf{Stage2.} Training student network. Firstly, we fix the teacher network and pre-train a 2x network to initialize student network. Then we use the jointly supervision loss to train student network. The initial learning rate is set to 5e-4 and halved at every 50 epochs and the total number of epochs is 500. The batch size and patch size are set to 64 and 256 separately. \\
\textbf{Stage3.} Fine-tune student network. (1) The student model is initialized from Stage2 and trained with the same settings as Stage2, especially the loss function is only MSE loss. (2) The student model is initialized from the previous step and fine-tuned by MSE loss further, it is worth that the patch size is set to 512. Other parameter settings are not changed.



\begin{figure*}[t]
    \centering
     \includegraphics[width=1\textwidth]{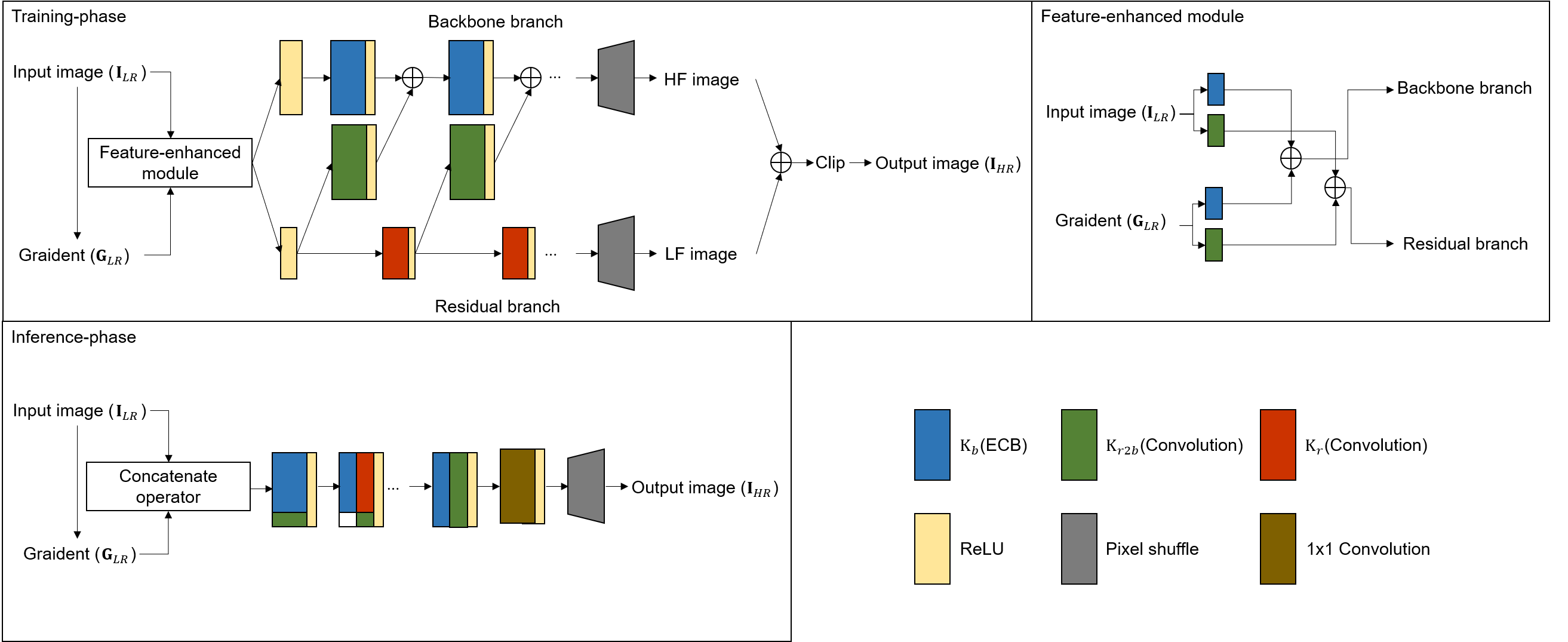}
    \caption{Overview of ETDS Network, conjugated with a Feature-Enhanced Module and an ECB Block -- Team MegastudyEdu.}
    \label{fig:megastudy}
\end{figure*}

\subsection{Enhancing RTSR with ETDS and Edge-oriented Convolutional Blocks.}
\label{sec:megastudy}

\begin{center}

\vspace{2mm}
\noindent\emph{\textbf{Team MegastudyEdu Vision AI}}
\vspace{2mm}

\noindent\emph{
Jae-Hyeon Lee,
Ui-Jin Choi
}

\vspace{2mm}

\noindent\emph{
MegastudyEdu Vision AI}

\end{center}


We introduce a method that leverages the Efficient Transformation and Dual Stream Network (ETDS) \cite{Chao_2023_CVPR} conjugated with a Feature-Enhanced Module and an Edge-oriented Convolution Block (ECB) \cite{zhang2021edge}. 
 

Our model is based on the Efficient Transformation and Dual Stream Network (ETDS) \cite{Chao_2023_CVPR}, incorporating a Feature-Enhanced Module inspired by Structure-Preserving Super Resolution with Gradient Guidance (SPSR)\cite{ma2020structure} and an Edge-oriented Convolution Block (ECB) proposed in ECBSR\cite{zhang2021edge}. This design utilizes the equivalent transformation to convert time-consuming operators into time-friendly operations, alongside a dual stream network structure to reduce redundant parameters.

The architecture of ETDS\cite{Chao_2023_CVPR} comprises Dual Stream network to alleviate redundant parameters, as follows:
\[\begin{bmatrix}K_b & K_{r2b} \\ K_{b2r} & K_r\end{bmatrix} \tag{1} \]

Where, $K_b$ (backbone branch) extracts high-frequency information, while $K_r$ (residual branch) processes low-frequency information. In our approach, ECB block is applied to $K_b$ to enhance efficiency, and $K_{r2b}$ and $K_r$ consist of 3x3 convolutions. 
Inspired by SPSR\cite{ma2020structure}, we add to restore information from images degraded by compression and downsampling algorithms. We extract gradient information from the input Low-Resolution (LR) images and then enhance the input feature map through a Feature-Enhanced Module. During inference, the Feature-Enhanced Module operates according to

\begin{equation*} z = \begin{bmatrix} W_1 \otimes x + b_1 \\ W_2 \otimes y + b_2 \end{bmatrix} = \begin{bmatrix} W_1 & O \\ O & W_2 \end{bmatrix} \otimes \begin{bmatrix} x \\ y \end{bmatrix} + \begin{bmatrix} b_1 \\ b_2 \end{bmatrix} \tag{2} \end{equation*}

transforming into a concatenate-convolution structure, with $K_b$ re-parameterized as a 3x3 convolution. Ultimately, all parameters are restructured through equivalent transformation, forming the comprehensive architecture of our model.\\
To confirm that our solution demonstrates superior performance over previous methods, we conducted a comparison between ETDS and our model. At AIS2024 CVPR, during the Validation phase for Real-Time Compressed Image Super-Resolution, it was observed that ETDS scored 22.844, whereas our proposed model scored 22.912, indicating an improvement in performance.

Our method is trained on DIV2K\cite{DIV2K} and Flickr2K datasets, with images processed using AVIF compression with Quality Factor (QF) coefficients ranging from 31 to 63, and scaled by a factor of 4 via Lanczos interpolation. During training we use data augmentation techniques:random cropping to 64x64, random flipping, and random rotation.

ETDS \cite{Chao_2023_CVPR} architecture is adapted with ECB \cite{zhang2021edge} to enhance edge detail recovery in high-frequency gradients, while the Feature-Enhanced Module, aids in restoring information lost through compression and downsampling.

\begin{table}[t]
    \centering
    \resizebox{\linewidth}{!}{
    \begin{tabular}{l c c c c c c}
        \toprule
        Model & PSNR & \# Params. (M) & FLOPs (G) & Runtime (ms)  \\
        \midrule
        ETDS \cite{Chao_2023_CVPR} & 22.844 & 0.0394 & 20.342 & 5.561 \\
        Our model & 22.912 & 0.0401 & 20.677 & 5.941 \\
        \bottomrule
    \end{tabular}
    }
    \caption{Ablation study by Team MegastudyEdu.}
    \label{tab:megastudy}
\end{table}

\paragraph{Implementation details}

\begin{itemize}
\item \textbf{Framework:} PyTorch 2.1.1, PyTorch Lightning
\item \textbf{Optimizer and Learning Rate:} We employed Adam optimizer with parameters $\beta_1 = 0.9$ and $\beta_2 = 0.999$. The training spanned 100 epochs with an initial learning rate set to 0.0001, halved at the 50th epoch.
\item \textbf{GPU:} NVIDIA A100 (80GB)
\item \textbf{Training Time:} The model trained for 24 hours.
\item \textbf{Training Strategies:} We trained the model using all AVIF images generated within the quality factor range of 31 to 63. This entailed training on a total of 110,400 images comprising 800 from DIV2K and 2,650 from Flickr2K, each at 32 quality factors.
\item \textbf{Efficiency Optimization Strategies:} 

\begin{itemize}
\item \textbf{Dual Stream Network Architecture:} Utilizing ETDS \cite{Chao_2023_CVPR} reduces redundant parameters by separating the processing of high-frequency and low-frequency information. This branch enables more efficient learning and reduces computational overhead.
\item \textbf{Feature-Enhanced Module with Gradient Guidance:} We incorporated a Feature-Enhanced Module to leverage gradient information from low-resolution inputs. This approach effectively restores high-frequency details lost during compression and downsampling, enhancing model performance without significantly increasing computational demand. \
\end{itemize}
\end{itemize}



\subsection{Unshuffle, Re-parameterization, and Pointwise Network (URPNet)}
\label{sec:URPNet}

\begin{center}

\vspace{2mm}
\noindent\emph{\textbf{Team 402Lab}}
\vspace{2mm}

\noindent\emph{
Hyeon-Cheol Moon~$^{1,2}$
Tae-Hyun Jeong~$^1$,
Yoonmo Yang~$^1$,
Jae-Gon Kim~$^2$,
Jinwoo Jeong~$^1$,
Sungjei Kim~$^1$
}

\vspace{2mm}

\noindent\emph{
$^1$ Korea Electronics Technology Institute (KETI)\\
$^2$ Korea Aerospace University (KAU)}

\end{center}


\begin{figure}[t]
\centering
\includegraphics[width=\linewidth]{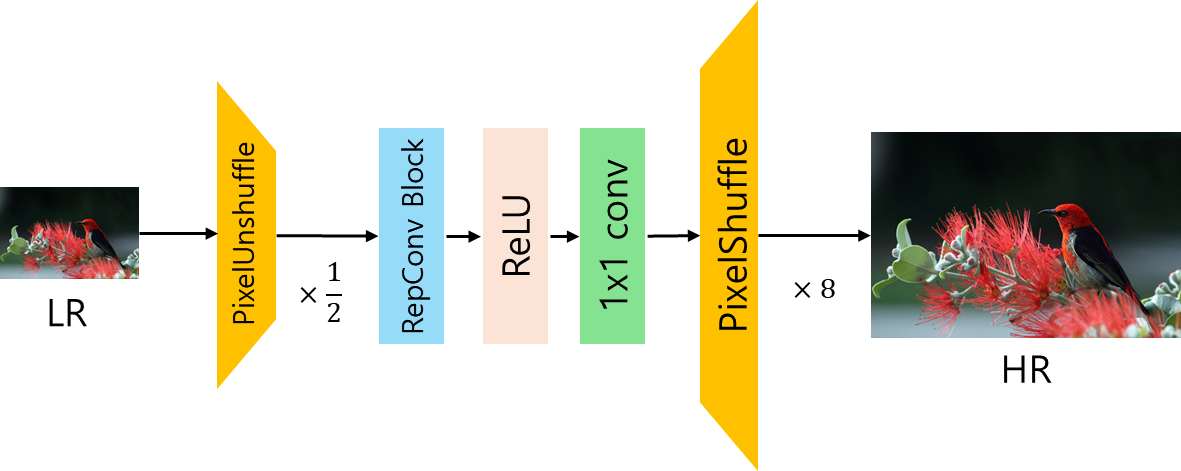}
\caption{Proposed URPNet structure.} 
\label{fig:URPNet}
\end{figure}

\begin{figure}[t]
\centering
\includegraphics[width=\linewidth]{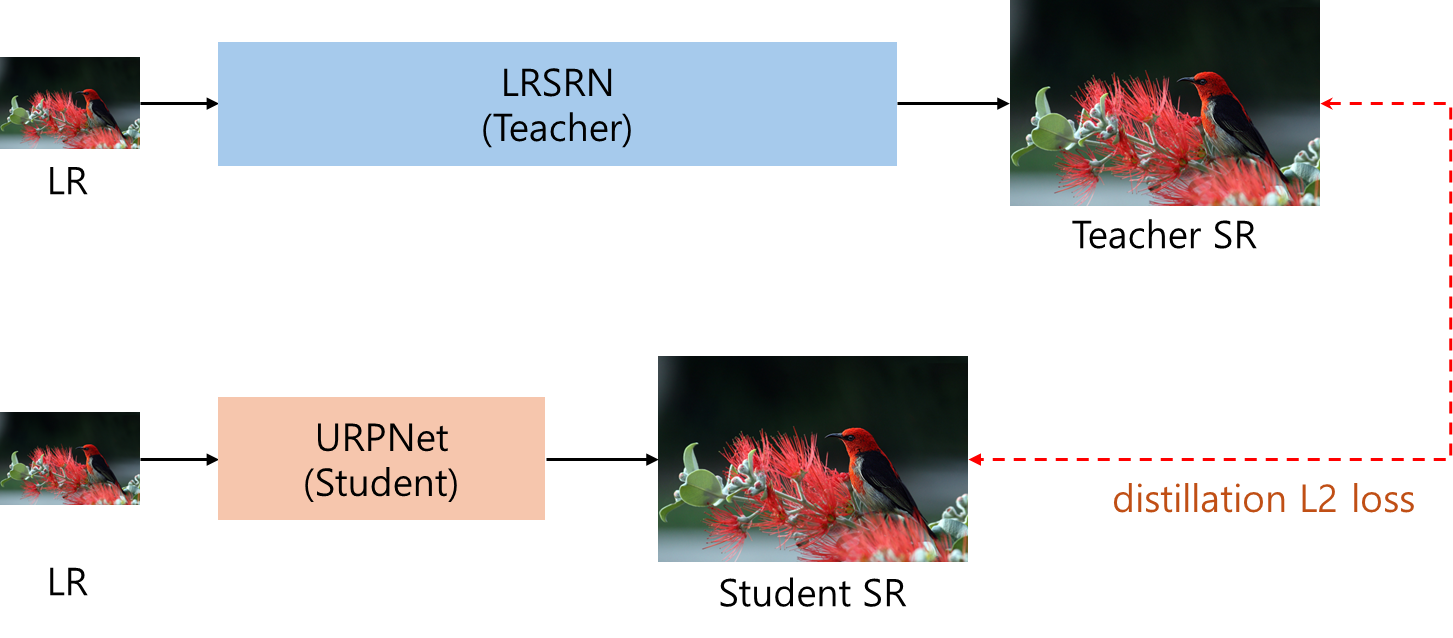}
\caption{Proposed distillation loss on the fine-tuning stage of URPNet, Team 402Lab.}
\end{figure}

We propose an Unshuffle, Re-parameterization, and Pointwise Network (URPNet) that can achieve higher accuracy at a faster speed compared to previous real-time SR models for 4K images. We applied a pixel unshuffle to the input image to reduce the resolution, and applied the 1x1 pointwise convolution to only the last layer, instead of applying a re-parameterized convolution (RepConv) to all existing convolutions.

We also applied curriculum learning \cite{curr} to efficiently learn lightweight models. Since the larger the Quantization Parameter (QP), the larger the compression artifacts, the worse the performance will be if the lightweight model is trained on high QP data from the beginning. Therefore, we divided the training data into easy (QP 31), medium (QP 39, QP 47), and hard (QP 55, QP 63) sets according to the training difficulty. 

Additionally, we applied knowledge distillation (KD) during the fine-tuning stage to achieve higher PSNR than using conventional training. To apply KD, the teacher model is trained from the scratch on a high-resolution dataset. We train with the L2 loss of the output images of each network between teacher and student \cite{facd, ICIP2020}. 

\paragraph{Efficiency metrics} Considering the challenge input, the model has 0.15K MACs per pixel (4K), a total number of 1.2483 GFlops, and a runtime of 0.62ms in RTX 3090 GPUs

\paragraph{Implementation details}

\begin{itemize}
    \item \textbf{Framework:} PyTorch 1.13 version 
    \item \textbf{Optimizer and Learning Rate:} Adam optimizer with a cosine warm-up. \\Initial learning rate: 5e-4 (scratch), 1e-4 (fine-tuning)
    \item \textbf{GPU:} single RTX3090/24GB, 3.2GB (training memory)
    \item \textbf{Datasets:} \begin{enumerate}
\item DIV2K : We use the DIV2K training dataset (800 images) for scratch training step.
\item FTCombined : We use a combined dataset for fine-tuning stage, which includes the DIV2K train set (full 800), Flickr train set (2650 full), DIV8K (first 200 samples), and LSDIR (first 1000). Before the training phase, the training data is pre-processed by center cropping it to a resolution of 2040 x 1080. To generate low-resolution images, we degrade the center cropped images with Lanczos downsampling and AVIF compression. For both training stages, we used random cropping, rotation 90, horizontal flip and vertical flip augmentation. 

\end{enumerate}
    \item \textbf{Training Time:} 24 hours with single RTX 3090GPUs
    \item \textbf{Training Strategies:} 
    \begin{enumerate}
    \item Scratch train step:
In the first step, our model was trained from scratch. The LR patches were cropped from LR images with 8 mini-batch 96x96 sizes. The Adam optimizer was used with a 0.0005 learning rate during scratch training. The cosine warm-up scheduler was used. The total number of epochs was set to 500. We use $l1$ loss.

\item Fine-tuning step:
In the second step, the model was initialized with the weights trained in the first step. To improve the accuracy, we used $l2$ and the distillation loss. Fine-tuning with $l2$ and distillation loss improves the peak signal-to-noise ratio (PSNR) value by 0.02 $\sim$ 0.03 dB. In this step, the initial learning rate was set as 0.0001, and the Adam optimizer was used along with a cosine warm-up. The total epoch was set to 50 epochs. 
\end{enumerate}

\end{itemize}


{\small
\bibliographystyle{ieee_fullname}
\bibliography{egbib}
}

\end{document}